\pdfoutput=1
\documentclass[11pt]{article}
\usepackage{acl}
\usepackage{times}
\usepackage{latexsym}
\usepackage[T1]{fontenc}
\usepackage[utf8]{inputenc}
\usepackage{microtype}
\usepackage{inconsolata}
\usepackage{array}
\usepackage{subcaption}

\usepackage{multirow}
\usepackage{booktabs}
\usepackage{graphicx}
\usepackage{soul}
\usepackage{pifont}

\newcommand{\reftbl}[1]{Table \ref{#1}}
\newcommand{\refsec}[1]{Section \ref{#1}}

\definecolor{highcolor}{RGB}{230, 255, 230}
\definecolor{mediumcolor}{RGB}{255, 255, 230}
\definecolor{lowcolor}{RGB}{255, 230, 230}

\newif\ifcomments
\ifcomments
    \newcommand{\todo}[1]{\textcolor{red}{[[ #1 ]]}\typeout{#1}}
    \newcommand{\harman}[1]{\textcolor{green}{HS: #1}}
    \newcommand{\shikhar}[1]{\textcolor{purple}{SB: #1}}
    \newcommand{\nitish}[1]{\textcolor{teal}{NG: #1}}
    \newcommand{\partha}[1]{\textcolor{brown}{PT: #1}}
\else
    \newcommand{\todo}[1]{}
    \newcommand{\harman}[1]{}
    \newcommand{\shikhar}[1]{}
    \newcommand{\nitish}[1]{}
    \newcommand{\partha}[1]{}
    
\fi

\newcommand{\checkmark}{\ding{51}}

\newcommand{\dataset}{\textsc{IndicGenBench}}
\newcommand{\xquad}{\textsc{XQuAD-In}}
\newcommand{\cssum}{\textsc{CrossSum-In}}
\newcommand{\flores}{\textsc{Flores-In}}
\newcommand{\xorqa}{\textsc{XorQA-In-En}}
\newcommand{\xorqaxx}{\textsc{XorQA-In-Xx}}
\newcommand{\xorqain}{\textsc{XorQA-In}}

\newcommand{\mt}{mT5}
\newcommand{\palm}{PaLM-2}
\newcommand{\llama}{LLaMA}
\newcommand{\gpt}[1]{{GPT-#1}}
\newcommand{\gptfam}{GPT}
\newcommand{\bloomz}{BLOOMZ}
\newcommand{\bloom}{BLOOM}
\newcommand{\gemma}{Gemma}
\newcommand{\airavata}{Airavata}

\newcommand\blfootnote[1]{
    \begingroup
    \renewcommand\thefootnote{}\footnote{#1}
    \addtocounter{footnote}{-1}
    \endgroup
}

\title{\dataset{}: A Multilingual Benchmark to Evaluate Generation Capabilities of LLMs on Indic Languages}

\author{
    Harman Singh,  
    \textbf{Nitish Gupta},
    \textbf{Shikhar Bharadwaj},
    \textbf{Dinesh Tewari}\textbf{,}
    \textbf{Partha Talukdar}
    \\
    Google Research India\\
    \texttt{\{hrman, guptanitish, shikharop, dineshtewari, partha\}@google.com} \\
}

\begin{document}
\maketitle

\begin{abstract}
As large language models (LLMs) see increasing adoption across the globe, it is imperative for LLMs to be representative of the linguistic diversity of the world. India is a linguistically diverse country of 1.4 Billion people. To facilitate research on multilingual LLM evaluation, we release \textit{\dataset{}} --- the largest benchmark for evaluating LLMs on user-facing generation tasks across a diverse set 29 of Indic languages covering 13 scripts and 4 language families.  
\dataset{} is composed of diverse generation tasks like cross-lingual summarization, machine translation, and cross-lingual question answering. %
\dataset{} extends existing benchmarks to many Indic languages through human curation providing multi-way parallel evaluation data for many under-represented Indic languages for the first time. We evaluate a wide range of proprietary and open-source LLMs including GPT-3.5, GPT-4, \palm{}, \mt{}, \gemma{}, \bloom{} and \llama{} on \dataset{} in a variety of settings. The largest \palm{} models performs the best on most tasks, however, there is a significant performance gap in all languages compared to English showing that further research is needed for the development of more inclusive multilingual language models. \dataset{} is available at \href{https://www.github.com/google-research-datasets/indic-gen-bench/}{www.github.com/google-research-datasets/indic-gen-bench} \blfootnote{All authors are now part of Google DeepMind}
\end{abstract}
\begin{table*}[t]
\Large
\centering
\resizebox{\textwidth}{!}{%
\begin{tabular}{l c m{16cm} m{4.3cm} c c}
\toprule

\multirow{2}{*}{\textbf{\Large{Task}}} & \multirow{2}{*}{\textbf{\Large{Language}}} & \multirow{2}{*}{\textbf{\Large{Input}}} & \multirow{2}{*}{\textbf{\Large{Output}}} & \textbf{\Large{\#Languages}} & \Large{\textbf{Dataset Size}} \\
& & & & \textbf{\small{(H / M / L)}} & \textbf{\small{(Train / Dev / Test)}} \\
\midrule
\shortstack[l]{\LARGE{\cssum{}} \\ (Cross-lingual Summarization)} & 
Hindi &
\includegraphics[scale=0.70,trim={0 0 0 0},clip]{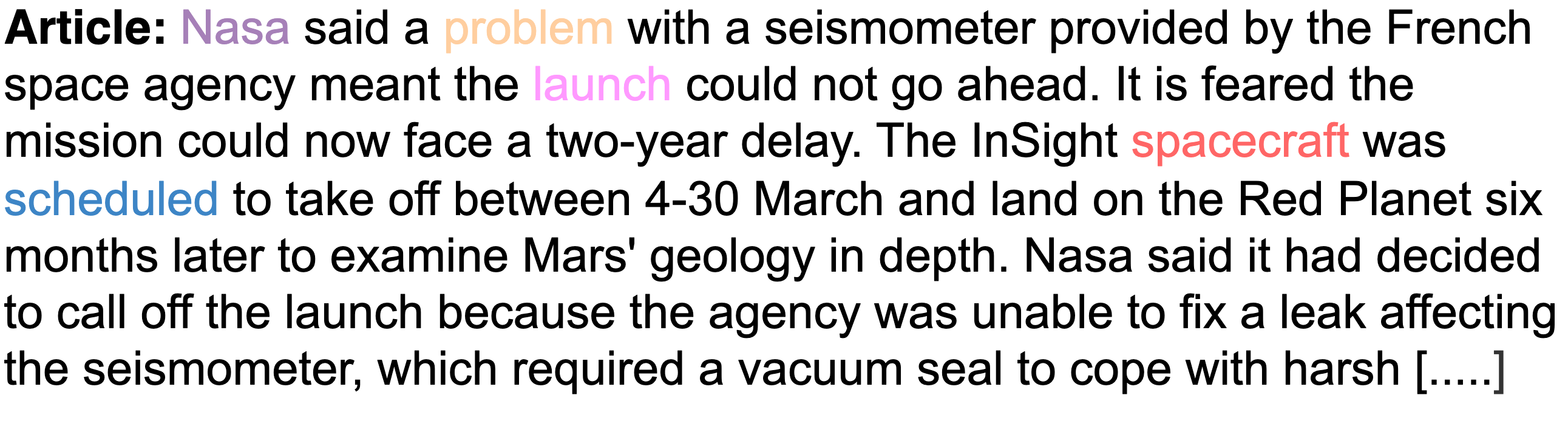} &
\includegraphics[scale=0.70,trim={0 0 0 0},clip]{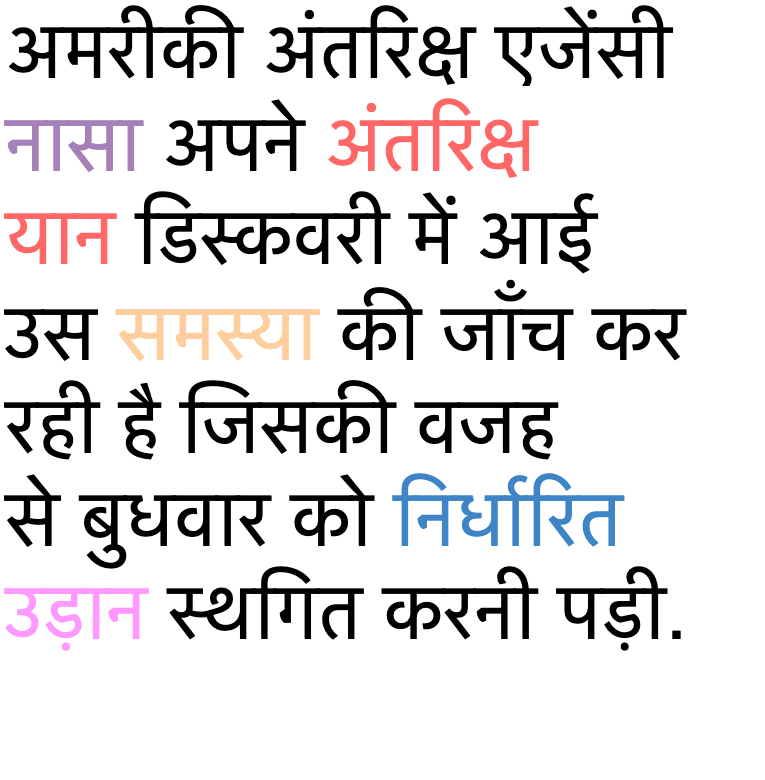} &
9 / 7 / 13 &
2.9k / 2.9k / 14.5k \\
\midrule
\shortstack[l]{\LARGE{\flores{}} \\ (Machine Translation)} & 
Konkani &
\includegraphics[scale=0.70,trim={0 0 0 0},clip]{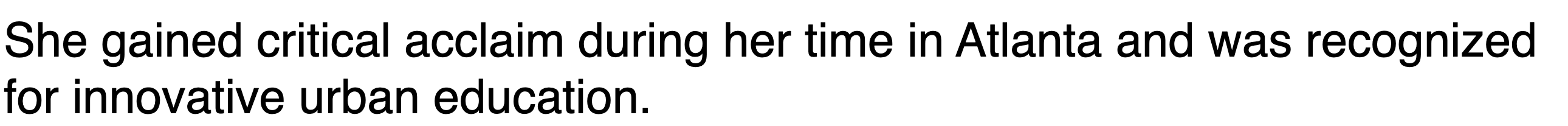} &
\includegraphics[scale=0.70,trim={0 0 0 0},clip]{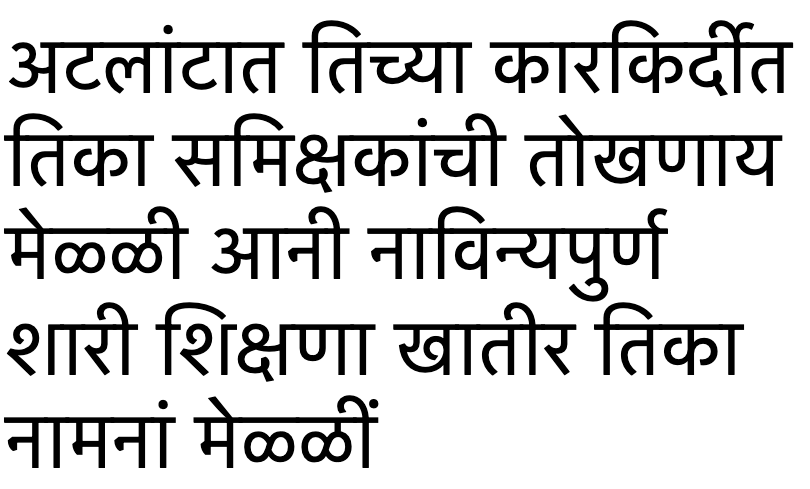} & 
9 / 7 / 13 & 
- / 28.9k / 29.3k \\
\midrule
\shortstack[l]{\LARGE{\xquad{}} \\ (Multilingual QA)} & 
Punjabi &
\includegraphics[scale=0.70,trim={0 0 0 0},clip]{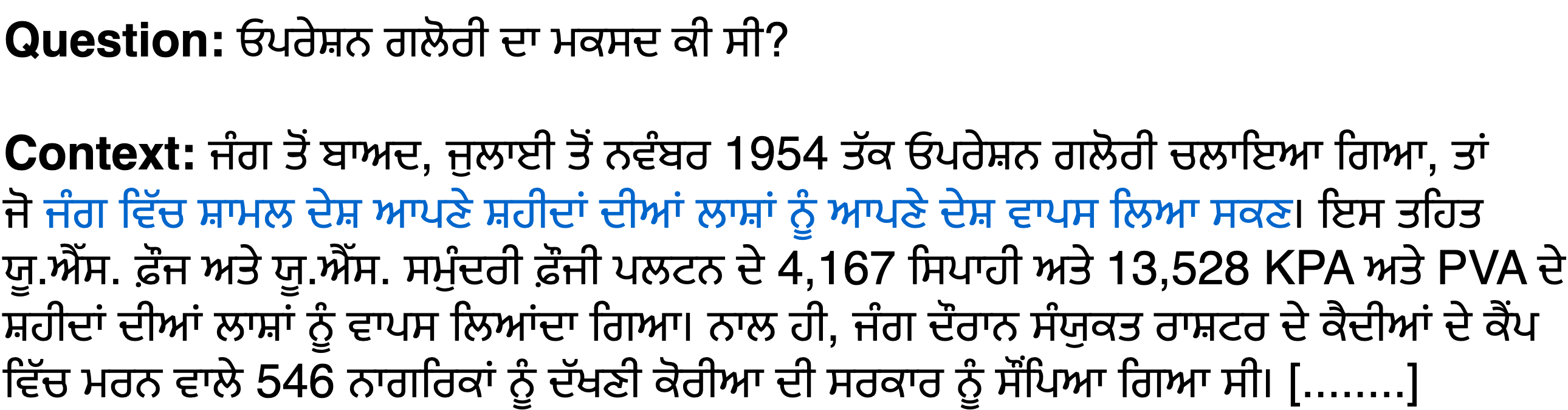} &
\includegraphics[scale=0.70,trim={0 0 0 0},clip]{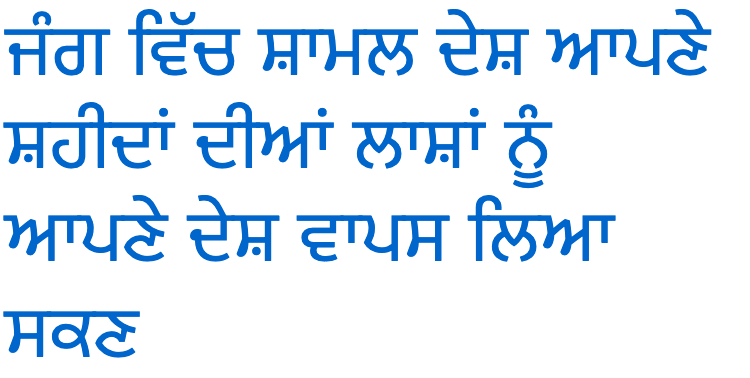} &
9 / 3 / - &
1.2k / 1.2k / 14.3k \\
\midrule
\shortstack[l]{\LARGE{\xorqaxx{}} \\ (Cross-lingual QA)} & 
Telugu &
\includegraphics[scale=0.70,trim={0 0 0 0},clip]{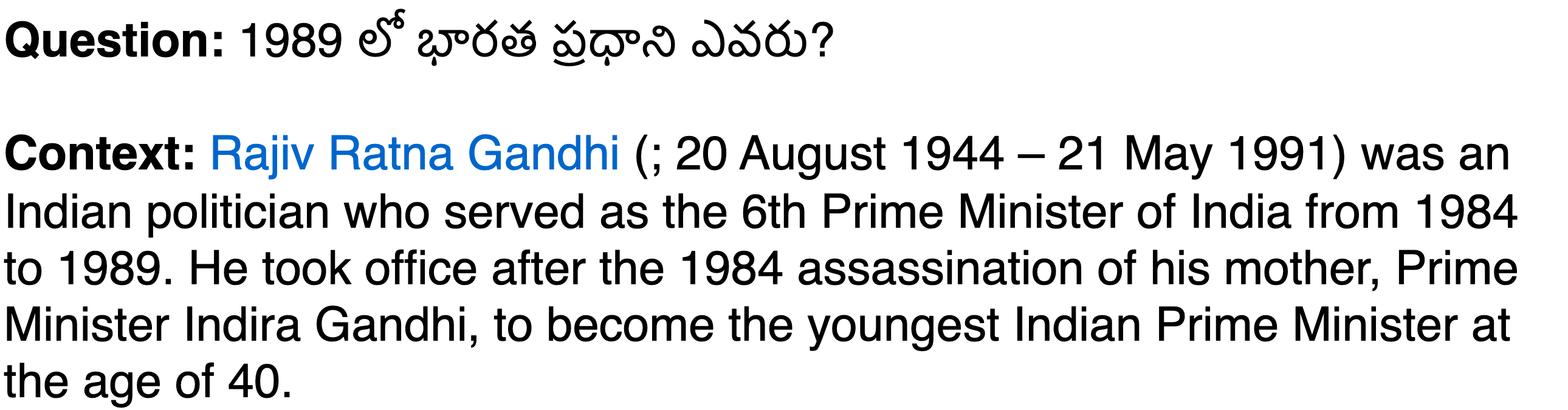} &
\includegraphics[scale=0.70,trim={0 0 0 0},clip]{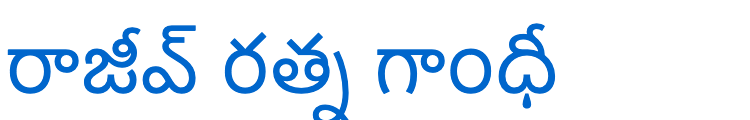} &
9 / 6 / 13 &
2.8k / 14k / 15.1k \\
\midrule
\shortstack[l]{\LARGE{\xorqa{}} \\ (Cross-lingual QA)} & 
Santali & 
\includegraphics[scale=0.70,trim={0 0 0 0},clip]{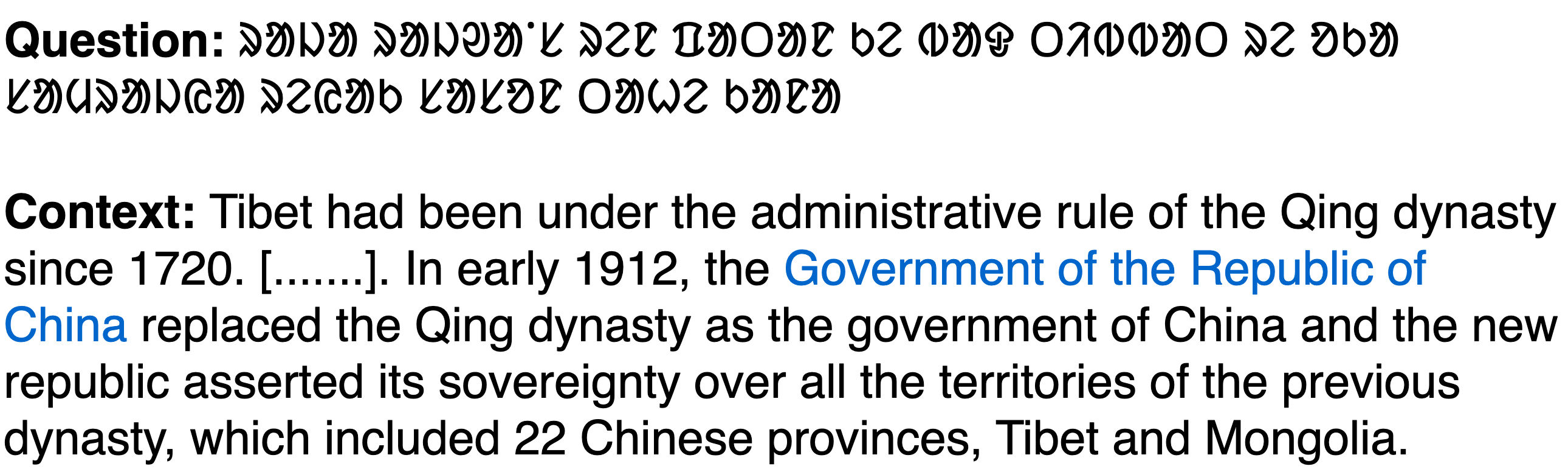} &
\includegraphics[scale=0.70,trim={0 0 0 0},clip]{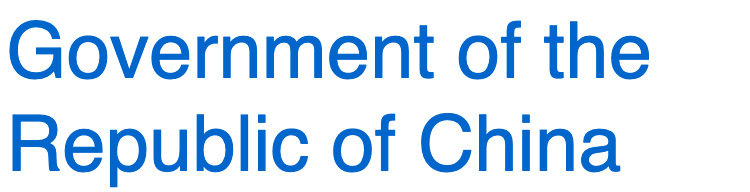} &
9 / 6 / 13 &
2.8k / 14k / 15.1k \\
\bottomrule
\end{tabular}
}
\caption{\dataset{}, our proposed benchmark, consists of five tasks: Cross-lingual Summarization (\cssum{}), Machine Translation (\flores{}), Multilingual QA (\xquad{}) and Cross-lingual QA (\xorqaxx{} and \xorqa{}). An example from each task, the number of languages for which we collect evaluation data (divided by resourcefulness, higher (H), medium (M) and low (L)), and the number of training/validation/test instances per task is shown above. See Section~\ref{sec:dataset_details} for details.}
\label{tab:task_examples}
\end{table*}

\begin{figure}[th!]
  \centering
  \includegraphics[width=\columnwidth]{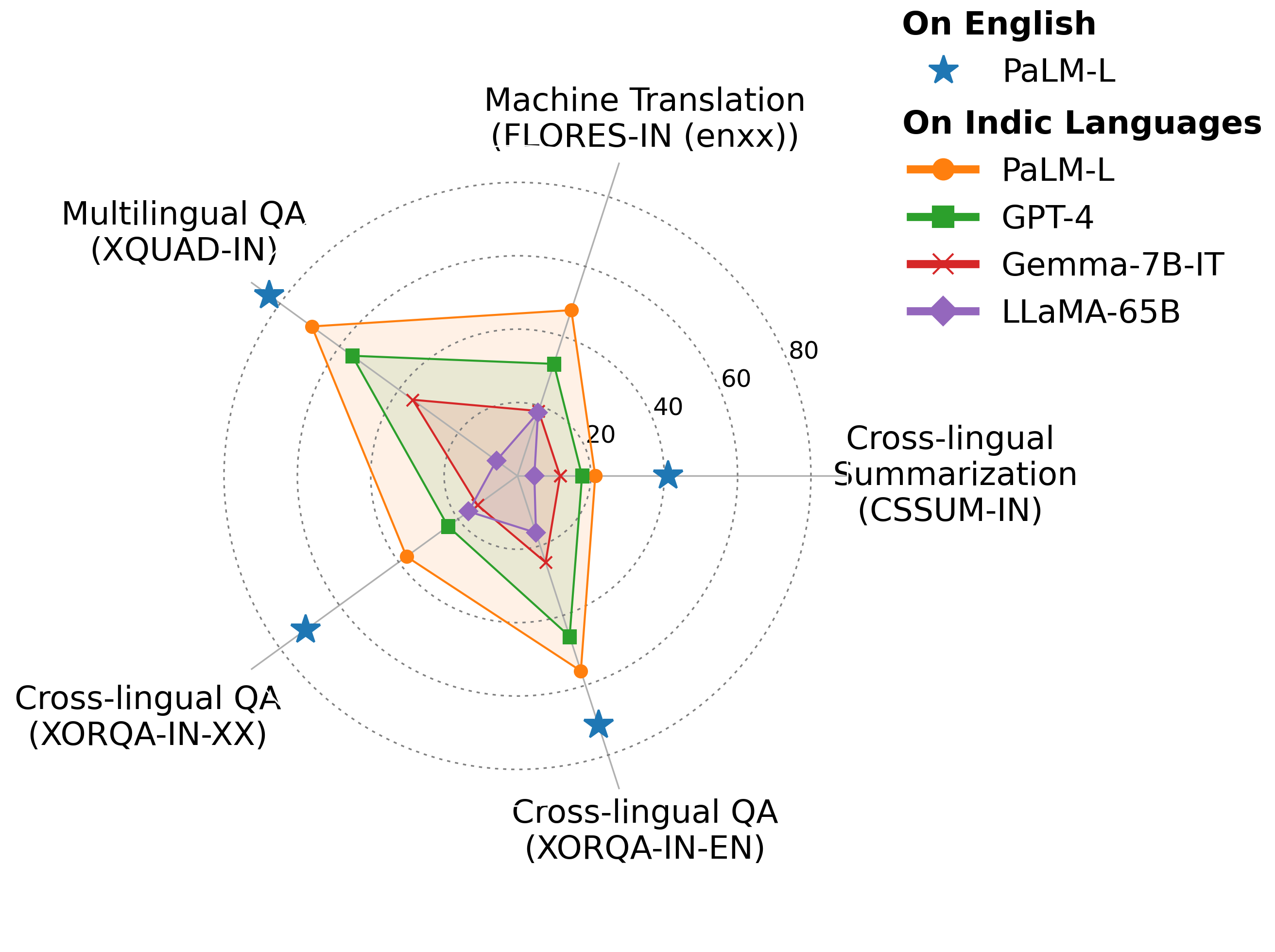}
  \caption{\textbf{Performance  of state-of-the-art LLMs on different tasks in \dataset{}}. We observe a significant performance gap between English and Indic languages across LLMs.
  }
  \label{fig:performance_radar_plot}
\end{figure}

\section{Introduction}
With the advances in generative language technologies powered by Large Language Models~\citep[LLMs;][]{gpt3, rae2022scaling, chowdhery2022palm, openai2023gpt4, tay2023ul2, Anil2023GeminiAF}, there has been a surge of interest in evaluating the multilingual capabilities of these models.
Recent work~\citep{ahuja2023mega, ahuja2023megaverse} shows a consistent performance gap between high resource languages and languages with lower amounts of web resources available.
To develop highly multilingual generative LLMs which should work equally well for 100s of languages spoken by billions of people in the world, it is crucial to evaluate their capabilities across a variety of languages to uncover performance gaps and guide future research.

In this work we focus on India, a country with 
1369 rationalized mother tongues spoken by more than a billion people.\footnote{\href{https://en.wikipedia.org/wiki/Languages\_of\_India}{https://en.wikipedia.org/wiki/Languages\_of\_India}} Making progress on language technologies for Indic languages will not only improve the state of affairs in this region, but will also provide valuable learning to the NLP community which will be applicable to other geographical regions and language families. 
There are has been much work from the community in building natural language \emph{understanding} (NLU) models for Indic languages~\citep{kakwani-etal-2020-indicnlpsuite, khanuja2021muril}, as well as evaluation datasets~\citep{doddapaneni-etal-2023-towards,mhaske-etal-2023-naamapadam} to support such models. In this work, our focus is to develop a high-quality benchmark for evaluating \emph{generative} language capabilities in a variety of Indic languages across various levels of resourcefulness. 

We release \dataset{}, a multilingual, multi-way parallel benchmark for measuring language generation capabilities across diverse user-facing tasks in 29 Indic languages across 4 language families (\reftbl{tab:lang_resource_based_classification}).
\dataset{} extends existing benchmarks such as CrossSum~\citep{bhattacharjee-etal-2023-crosssum}, XQuAD~\citep{xquad}, XorQA~\citep{xorqa}, and FLORES~\citep{nllbteam2022language} for additional Indic languages and is composed of tasks like cross-lingual summarization (\cssum{}), machine translation (\flores{}), cross-lingual reading comprehension (\xorqaxx{} and \xorqa{}) and multilingual reading comprehension (\xquad{}).
Each dataset consists of parallel examples in up to 29 low to comparatively higher resource Indic languages; and for some tasks (e.g. \cssum{}), \dataset{} provides the first-ever evaluation set for as many as 18 of these languages.
We also release a small training set in all tasks for efficient adaptation of LLMs. 
Our comprehensive evaluation of various state-of-the-art proprietary and open-source LLMs on \dataset{} shows that there is a significant gap in performance between English and Indic languages (see Figure \ref{fig:performance_radar_plot}).
Our contributions are as follows:
\begin{itemize}
    \item Created and released \dataset{}, a high quality text benchmark in diverse language generation tasks like summarization, question-answering, and translation across 29 Indic languages. \dataset{} is the largest generation benchmark for Indic languages.

    \item Comprehensive experimentation on SoTA LLMs (\mt{}, \gemma{}, \bloom{}, \llama{}, \gpt{3.5}, \gpt{4}, \palm{}) across various model sizes and training settings to benchmark their Indic language generation capabilities.

    \item A qualitative analysis for assessing the gaps in current language technologies and define potential directions of future research.
\end{itemize}
\section{\dataset{}}
\label{sec:dataset_details}

\dataset{} is a high-quality, human-curated benchmark to evaluate text generation capabilities of multilingual models on Indic languages. 
Our benchmark consists of 5 user-facing tasks (viz., summarization, machine translation, and question answering) across 29 Indic languages spanning 13 writing scripts and 4 language families. For certain tasks, \dataset{} provides the first-ever evaluation dataset for up to 18 Indic languages.
\reftbl{tab:task_examples} provides summary of \dataset{} and examples of instances across tasks present in it.

Languages in \dataset{} are divided into (relatively) Higher, Medium, and Low resource categories based on the availability of web text resources (see appendix \S\ref{app:lang_resource_classification} for details).\footnote{We note that the languages called relatively higher resource in this paper,  e.g., Hindi or Bengali, are in fact mid-low Web resource when compared to English and other truly high resource languages. For example, using Wikipedia as a proxy for language resources, compared to 6.6M+ Wikipedia articles in English, there are only 160K Hindi Wikipedia articles.}
\\

\noindent\fbox{\begin{minipage}{0.95\columnwidth}
\noindent \textbf{Higher} (9): {\small Bengali, Gujarati, Hindi, Kannada, Malayalam, Marathi, Tamil, Telugu, Urdu} \\
\noindent \textbf{Medium} (7): {\small Assamese, Bhojpuri, Nepali, Odia, Punjabi, Pashto, Sanskrit} \\
\noindent \textbf{Low} (13): {\small Awadhi, Haryanvi, Tibetan, Garhwali, Konkani, Chhattisgarhi, Rajasthani, Maithili, Manipuri, Malvi, Marwari, Santali, Bodo} 
\end{minipage}
}
\vspace{0.5mm}
\\
As evident from the lists above, our benchmark provides a broad-coverage over languages with respect to their resourcedness, allowing users to evaluate language models on relatively high-resource languages such as Hindi and extremely low-resource languages such as Manipuri in Meitei script on a single benchmark.

To curate the evaluation datasets for our benchmark, we use the following existing datasets as the source:
CrossSum~\cite{bhattacharjee-etal-2023-crosssum} for cross-lingual summarization, FLORES~\cite{nllbteam2022language} for machine translation, 
XQuAD~\cite{xquad} for multilingual QA, and
XoRQA~\cite{xorqa} for cross-lingual QA.
From each of these datasets we select a subset of English examples to be a part of our benchmark, and then collect professional human translations for these examples in all target Indic languages. 
Some target languages are already covered by the source datasets in which case we re-purpose this existing data and only collect translations for the remaining languages. We also collect and release a small amount of training and validation data making possible evaluation of training techniques like fine-tuning, parameter-efficient training, in-context learning, and others.

\noindent \textbf{Why extend existing benchmarks?} We chose to collect human translations of existing benchmarks as opposed to creating evaluation data from scratch due to various reasons:
\begin{itemize}
    \item Translation-based extension of existing benchmark results in multi-way parallel data, allowing researchers to attribute performance due to task knowledge vs. language understanding, and measure cross-lingual generalization
    \item For many low-resource languages in \dataset{}, clean text knowledge corpus (e.g., Wikipedia) is not available making it difficult to acquire source data for annotation
    \item By focusing only on translation quality in the target Indic languages, we are able to leverage the quality control that went into designing the source benchmarks.
\end{itemize}

\noindent Annotators were professional data labelers working as contractors at our organization and with a vendor. Annotators were paid competitive rates in compliance with applicable labor laws and prevailing market rates. Our pay rate to annotators varied across languages, ranging from USD 2.80 per hour for Pashto to USD 15.90 per hour for Tibetan. 

\subsubsection*{Cross-Lingual Summarization: \cssum{}}
We create \cssum{} based on CrossSum~\cite{bhattacharjee-etal-2023-crosssum}, a dataset for cross-lingual summarization, which in turn is derived from XL-Sum \citep{Hasan2021XLSumLM}. CrossSum contains multi-way parallel data in 45 languages where BBC news articles as source in a language are paired with corresponding summaries in other languages. Based on their matching criteria, different languages have different amount of source-target pairs. 

We sample 700 English article-summary pairs (100 each from train/dev and 500 from test) and ask human translators to translate the English summary into the target Indic languages. CrossSum already contains data for 9 of our 29 target languages; for these languages we sample 100/100/500 examples from the original dataset to maintain equity with other languages we collect data for. \cssum{} contains a total of 20.3k examples across 29 Indic languages in our benchmark.

\subsubsection*{Machine Translation: \flores{}}
FLORES-200~\citep{nllbteam2022language} is a human-annotated multi-way parallel machine translation (MT) benchmark for 200 languages where the same source English sentences are translated by humans into the target 200 languages. It contains data in 22 of our 29 target languages; we extend this by collecting human translations for the remaining 7 new languages leading to a MT benchmark in 29 Indic languages which we call \flores{}.

FLORES-200 is divided into three splits: dev (997), devtest (1012), test (992), of which the test set it not public. We collect translations for all 997 dev and 1012 devtest sentences, yielding 2009 sentences per language. Collectively, \flores{} contains 58.2k examples across 29 Indic languages.

\subsubsection*{Multilingual Question-Answering: \xquad{}} 
We create an Indic Multilingual Question Answering task \xquad{} based on the multilingual reading comprehension dataset XQuAD~\citep{xquad}. XQuAD is in turn derived from the SQuAD dataset~\citep{Rajpurkar2016SQuAD1Q}, in which an English Wikipedia passage is paired with multiple question-answer (QA) pairs where the answers are short spans for the given passage.
The authors of XQuAD collected human translations for 240 passages and 1190 QA pairs from the SQuAD v1.1 development set into 10 higher resource languages (Hindi being the only Indic language). 

To create \xquad{}, we use the 240 passages and 1190 QA pairs from XQuAD as our test set. 
We additionally selected 20 passages and 100 QA pairs from the original SQuAD v1.1 training and development sets each to create our training and development set. 
For all the 280 passages and 1390 QA pairs we collect professional human translations in 12 Indic languages.\footnote{\xquad{} contains all 9 higher-resource languages (see \S\ref{sec:dataset_details}) and 3 medium-resources languages, namely, Assamese, Odia, and Punjabi.}
Overall, \xquad{} contains 3.3k passages and 16.6k QA pairs in 12 Indic languages.

\subsubsection*{Cross-Lingual Question-Answering: \xorqain{}}
We create Indic Cross-lingual Question-Answering dataset \xorqain{} based on the \textsc{Xor-TyDi} QA dataset~\citep{xorqa}. \textsc{Xor-TyDi} contains questions in non-English languages paired with English evidence passages and short span answers from those passages (similar to SQuAD). It was created with the idea of developing NLP systems that can answer questions in users' native language by refering to sources in a high-resource language, such as English, which was more likely to contain the answer due to the \textit{information scarcity} of low-resources languages on the web. The original \textsc{Xor-TyDi} contains data in 7 languages out of which Bengali and Telugu are the two Indic languages.

To create \xorqain{}, we select the 302 Bengali and 237 Telugu examples (Bn/Te-question, En-passage, En-answer) from the \textsc{Xor-TyDi} dev set as our test data.\footnote{\textsc{Xor-TyDi} has not publicly released its test set.}
Additionally, we sample 600 examples (equally from Bengali and Telugu) from the training set of \textsc{Xor-TyDi} to create our training (100) and development (500) set. 
We then follow a two-staged translation process, where we first ask the human translators to translate the Bengali or Telugu question (Bn/Te-question) into English (En-question). 
In the second stage, we collect translations for these English questions (En-question) into target languages (Xx-question) and translations for the English answers (En-answer) into the target languages (Xx-answer). 

We create two tasks from this translated data:
1. {\textbf \xorqa{}}: Each example contains (Xx-question, En-passage, En-answer). This task is similar to the \textsc{Xor-TyDi} dataset. \\
2. {\textbf \xorqaxx{}}: Each example contains (Xx-question, En-passage, Xx-answer), where the task is to generate the answer in the same language as the question. 

We collect data for 28 Indic languages resulting in 32k examples.\footnote{We do not collect translations for Nepali.}

See Appendix Table \ref{tab:lang_coverage} for languages covered by each dataset in \dataset{}.

\begin{table}[t]
\fontsize{7}{10pt}\selectfont
  \centering
  \begin{tabular}{@{}l@{} c@{}  c@{}  c@{}  c@{}  c@{}}
    \toprule
    \textbf{Model (LLM)} &
    \multicolumn{1}{c}{\textbf{\rotatebox[origin=c]{63.5}{\tiny{\cssum{}}}}} &
    \multicolumn{1}{c}{\textbf{\rotatebox[origin=c]{63.5}{\tiny{\flores{}}}}} &
    \multicolumn{1}{c}{\textbf{\rotatebox[origin=c]{63.5}{\tiny{\xquad{}}}}} & 
    \multicolumn{1}{c}{\textbf{\rotatebox[origin=c]{63.5}{\tiny{\xorqaxx{}}}}} &
    \multicolumn{1}{c}{\textbf{\rotatebox[origin=c]{63.5}{\tiny{\xorqa{}}}}}  \\
    \textbf{Eval. Metric} & ChrF & ChrF & Token-F1 & Token-F1 & Token-F1 \\
     &  & \texttt{(enxx / xxen)} & & & \\
    
    \midrule
    \multicolumn{6}{@{}l}{\fontsize{8}{10pt}\selectfont \textit{Performance in \textbf{English}}} \\
    \midrule
    
    GPT-4      & 30.3 & -- / -- & 64.8  & -- & 37.9 \\
    \palm{}-L  & 41.1 & -- / -- & 83.7  & --  & 71.4 \\
    
    \midrule
    \multicolumn{6}{@{}l}{\fontsize{8}{10pt}\selectfont \textit{Average Performance on \textbf{\dataset{}}}} \\
    \midrule
    \llama{}-7B  & 3.7 & 11.5 / 21.6 & 3.8 & 7.4  & 10.4  \\
    \llama{}-13B & 4.1 & 13.3 / 24.1 & 4.5 & 10.4 & 12.1  \\
    \llama{}-65B & 4.6 & 18.1 / 32.7 & 7.1 & 16.5 & 16.3  \\
    \addlinespace[1mm]
    \bloom{}-7B & 3.8 & 18.3 / 31.2 & 13.8 & 7.9 & 23.6 \\
    \bloomz{}-7B & 1.2 & 40.8 / 48.4 & 53.7 & 7.0 & 49.0 \\
    \addlinespace[1mm]
    \gemma{}-7B-PT & 0.0 & 32.1 / 50.4 & 0.5 & 11.7 & 23.8 \\
    \gemma{}-7B-IT & 11.6 & 18.6 / 29.2 & 35.3 & 13.5 & 24.8 \\
    \addlinespace[1mm]
    GPT-3.5 & 16.3 & 29.2 / 47.7 & 33.2 & 21.6 & 35.5 \\
    GPT-4   & 17.6 & 32.1 / 54.5 & 55.7 & 23.4 & 46.0 \\
    \addlinespace[1mm]
    \palm{}-XXS & 7.2  & 24.0 / 43.4 & 34.6 & 13.5  & 36.8 \\
    \palm{}-XS  & 15.5 & 40.7 / 58.3 & 62.2  & 29.5  & 47.8\\
    \palm{}-S   & 18.5  & 43.5 / 61.6 & 66.7 & 31.6  & \textbf{57.4}\\
    \palm{}-L  & \textbf{21.2} & \textbf{47.5 / 65.1} & \textbf{69.3}  & \textbf{37.4}  & 55.9 \\
    \bottomrule
  \end{tabular}
  \caption{\label{tab:one-shot-results-avg}\textbf{One-shot performance on \dataset{}} across model sizes for all LLMs considered in our work~(\S\ref{ssec:one-shot-performance}). For each LLM family performance improves with increasing model size, with \palm{}-L performing the best across most tasks. Compared to English, all models under-perform significantly highlighting shortcomings of current SoTA LLMs.
  See Section \ref{ssec:one-shot-performance} for details.
  }
  \vspace{-0.5cm}
\end{table}

\begin{table*}[ht!]
\fontsize{6}{10pt}\selectfont
  \centering
  \begin{tabular}{@{\hspace{0cm}}l@{\hspace{0.3cm}}ccc|ccc|cc|ccc|ccc}
    \toprule
    \multicolumn{1}{l}{} &
    \multicolumn{3}{c}{\textbf{\cssum{}}} & 
    \multicolumn{3}{c}{\textbf{\flores{}} (\texttt{enxx} / \texttt{xxen})} &
    \multicolumn{2}{c}{\textbf{\xquad{}}} & 
    \multicolumn{3}{c}{\textbf{\xorqaxx{}}} & 
    \multicolumn{3}{c}{\textbf{\xorqa{}}}
     \\
    \cmidrule(lr){2-4} \cmidrule(lr){5-7} \cmidrule(lr){8-9} \cmidrule(lr){10-12} \cmidrule(lr){13-15}
    \textbf{Model}

    & \textbf{High}  & \textbf{Medium}  & \textbf{Low} 
    & \textbf{High}  & \textbf{Medium}  & \textbf{Low} 
    & \textbf{High}  & \textbf{Medium} 
    & \textbf{High}  & \textbf{Medium}  & \textbf{Low} 
    & \textbf{High}  & \textbf{Medium}  & \textbf{Low}  \\
    \midrule
    \llama{}-65B & 4.4 & 4.6 & 4.7 & 18.2 / 31.5 & 15.4 / 30.0 & 19.5 / 35.0 & 8.8 & 1.9 & 17.7 & 13.5 & 17.1 & 16.4 & 14.0 & 17.3\\

    \gemma{}-7B-IT &  13.9 & 11.5 & 10.0  & 17.6 / 33.7 & 15.0 / 26.1 & 21.3 / 27.7 & 38.8 & 24.8 &  18.9 & 8.3 & 12.2  &  29.5 & 23.9 & 21.9  \\
    \bloomz{}-7B &  1.5 & 1.7 & 0.6  & \textbf{67.7} / 59.1 & 39.4 / 50.2 & 22.9 / 40.0 & 55.5 & 48.1 &  10.8 & 2.8 & 6.2  &  \textbf{64.7} & 45.8 & 39.5\\

    \gpt{4}  
    & 19.4 & 17.9 & 16.3
    & 36.2 / 59.6 & 30.7 / 55.2 & 29.9 / 50.5
    & 56.1 & 54.6 
    & 25.8 & 21.6 & 22.6 
    & 49.4 & 50.0 & 41.8 \\
    \palm{}-L 
    & \textbf{25.2} & \textbf{23.1} & \textbf{17.5}
    & 56.9 / \textbf{68.2} & \textbf{45.9 / 65.6} & \textbf{41.9 / 62.6}
    & \textbf{72.5} & \textbf{59.8} 
    & \textbf{41.9} & \textbf{36.7} & \textbf{34.6} 
    & 57.3 & \textbf{57.9} & \textbf{53.9} \\
    
    \bottomrule
  \end{tabular}
  \caption{\textbf{One-shot performance across language categories} based on resourcedness defined in Section~\ref{sec:dataset_details}. For all tasks, we witness significantly lower performances in medium and low resource languages compared to the higher resource ones.
  Please see Table~\ref{tab:one-shot-results-lang-resource-allmodels} in appendix~\ref{app:lang-resources-allmodels} for results on other models. See Section \ref{ssec:results-across-lang-resource} for more details.
}
  \label{tab:one-shot-results-lang-resource}
\end{table*}

\section{Experiments and Analysis}
We use \dataset{} to benchmark multilingual and cross-lingual language generation capabilities of various LLMs on Indic languages.
We perform experiments with a variety of 
open-source LLMs --- \mt{}~\citep{xue-etal-2021-mt5}, \llama{}~\citep{touvron2023llama},\footnote{\llama{}-2 could not be used due to a restrictive licence}, \bloomz{}~\citep{workshop2022bloom}, \gemma{}~\citep{gemmateam2024gemma};
and proprietary LLMs --- \gpt{3.5}, \gpt{4}~\citep{openai2023gpt4}, and \palm{}~\citep{anil2023palm}.

We compare and analyze the performance of different model size variants of these LLMs under various learning paradigm settings. We first evaluate model performance on one-shot prompting~(\S\ref{ssec:one-shot-performance}) and also measure performance across language categories based on resourcedness~(\S\ref{ssec:results-across-lang-resource}). We then evaluate the effect of number of in-context examples shown to the model as supervised data~(\S\ref{ssec:in-context-learning}) and the effect of prompting in a higher-resource language such an English or Hindi~(\S\ref{ssec:transfer-learning}). Using the training data contained in \dataset{}, we measure how the performance of LLMs after fine-tuning compares with few-shot prompting~(\S\ref{ssec:fine-tuning}). Finally, we perform qualitative analysis of models on \dataset{} and highlight some areas of improvement for future model development~(\S\ref{ssec:qualitative-analysis}).

\paragraph{Evaluation Metrics} For the cross-lingual summarization and translation tasks, \cssum{} and \flores{}, we report Character-F1 (ChrF) metric~\citep{popovic-2015-chrf} since token-level metrics like ROUGE and BLEU are not reliable for low-resource languages~\citep{Bapna2022BuildingMT}. To stay consistent with existing literature on QA tasks, we report SQuAD-style Token-F1 on our \xquad{} and \xorqain{} QA tasks.\\
On \flores{}, we report translation performance in both directions---translating from English to the target language (\texttt{enxx}) and vice-versa (\texttt{xxen}).

\subsection{Comparison of LLMs on \dataset{}}
\label{ssec:one-shot-performance}

In Table~\ref{tab:one-shot-results-avg} we evaluate \llama{}, \bloomz{}, \gemma{}, \gptfam{} and \palm{} family of models on all tasks of \dataset{} in a one-shot prompted setting. Numbers are averaged across all languages in the evaluation data. To compare, we also report English performance for \gpt{4} and \palm{}-L. 

We see across tasks that larger models from the same LLM family perform better.
\palm{}-L performs the best among all LLMs considered, except for the \xorqa{} task where \palm{}-S performs slightly better.
We find that open source \llama{} models perform much worse compared to proprietary models; even the largest \llama{}-65B model significantly underperforms the smallest \palm{}-XXS model. \gemma{}-7B instruction tuned model performs better than \llama{}-13B as well as \llama{}-65B on most tasks. \bloomz{}, which is an instruction tuned version of \bloom{}~\citep{workshop2022bloom}, pre-trained on large-scale multilingual data, works the best on three out of five tasks in \dataset{}. On \cssum{} and \xorqaxx{} it falls behind \llama{} and \gemma{}.
Compared to English, we see significant room for improvement (20+ ChrF or Token-F1 points) across all tasks.

\subsection{Performance across language categories}
\label{ssec:results-across-lang-resource}
In Table~\ref{tab:one-shot-results-lang-resource} we report one-shot performance across language categories defined in Section~\ref{sec:dataset_details}. We only show performance for \gemma{}-7B-IT, \bloomz{}-7B, \llama{}-65B, \gpt{4} and \palm{}-L models here and report performance for the other models in appendix~\ref{app:lang-resources-allmodels}. We find that there is a significant performance drop going from higher resourced languages to medium resourced ones, and further drop in lower resourced languages.

\noindent
We would like to point out two observations here: \\
(a) In \flores{}, the performance for translating English to the target language (\texttt{enxx}) drops significantly from higher to lower resourced languages ($56.9 \rightarrow 41.9$ for \palm{}-L) whereas the performance in the \texttt{xxen} direction does not fall this drastically ($68.2 \rightarrow 62.6$). This is also seen in \xorqaxx{} and \xorqa{}. This highlights that current LLMs are better at understanding than generation in these lower-resourced languages. \\
(b) In few cases, we see smaller performance deltas between medium and lower resourced languages compared to higher and medium categories. From our analysis, this can mainly be attributed to many languages in the lower category being similar to Hindi and written in the same Devanagari script.

\begin{table}[ht!]
\fontsize{7}{10pt}\selectfont
  \centering
  \begin{tabular}{lccc|cccc}
    \toprule
     & \multicolumn{3}{c}{\textbf{\flores{}}} & \multicolumn{4}{c}{\textbf{\xorqaxx{}}} \\
    \cmidrule(lr){2-4} \cmidrule(lr){5-8}
    \textbf{Model (LLM)} & \textbf{0} & \textbf{1} & \textbf{5} & \textbf{0} & \textbf{1} & \textbf{2} & \textbf{3} \\
    \midrule
    \llama{}-7B & 8.0 & 11.5 & 11.4 & 5.0 & 7.4 & 9.0 & 9.2 \\
    \llama{}-13B & 8.6 & 13.3 & 13.4 & 6.3 & 10.4 & 12.2 & 13.1 \\
    \llama{}-65B & 14.0 & 18.1 & 18.3 & 12.3 & 16.5 & 18.7 & 19.4 \\
    \midrule
    \palm{}-XXS & 0.8 & 24.0  & 26.9 & 8.9 & 13.5 & 15.8 & 17.5 \\
    \palm{}-XS & 20.1 & 40.7 & 42.3 & 21.4 & 29.5 & 32.2 & 33.2 \\
    \palm{}-S & 24.9 & 43.5 & 45.2 & 22.7 & 31.6 & 33.4 & 35.4 \\
    \palm{}-L & \textbf{31.1} & \textbf{47.5} & \textbf{49.3} & \textbf{31.9} & \textbf{37.4} & \textbf{39.7} & \textbf{41.1} \\
    \bottomrule
  \end{tabular}
  \caption{\textbf{Performance by varying number of in-context exemplars} for \llama{} and \palm{} models on \flores{} (\texttt{enxx}) and \xorqaxx{} tasks~(\S\ref{ssec:in-context-learning}). Performance improves with increasing amounts of supervision provided in-context. Refer appendix~\ref{app:few-shot-complete} for results on other tasks and models.}
  \label{tab:few_shot_flores_avg}
\end{table}

\begin{table*}[ht!]
\fontsize{7.5}{10pt}\selectfont
  \centering
  \begin{tabular}{lccc|cc|ccc|ccc}
    \toprule
    \multicolumn{1}{l}{} 
    & \multicolumn{3}{c}{\textbf{\cssum{}}} 
    & \multicolumn{2}{c}{\textbf{\xquad{}}} 
    & \multicolumn{3}{c}{\textbf{\xorqaxx{}}} 
    & \multicolumn{3}{c}{\textbf{\xorqa{}}} \\
    \cmidrule(lr){2-4} \cmidrule(lr){5-6} \cmidrule(lr){7-9} \cmidrule(lr){10-12}
    \multicolumn{1}{l}{\textbf{Model (1-Shot Lang)}} & \textbf{Higher} & \textbf{Medium} & \textbf{Low} & \textbf{Higher} & \textbf{Medium} & \textbf{Higher} & \textbf{Medium} & \textbf{Low} & \textbf{Higher} & \textbf{Medium} & \textbf{Low}\\
    \midrule
    
    \palm{}-XXS (En) &
    0.3 & 0.1 & 0.3 &
    \underline{38.5} & \underline{31.9} &
    14.0 & 5.4 & 7.3 &
    40.3 & 35.0 & 30.8 \\
    
    \palm{}-XXS (Hi) &
    \underline{1.3} & \underline{2.1} & \underline{3.7} &
    \textbf{39.8} & \textbf{33.3} &
    \underline{17.6} & \underline{8.5} & \underline{10.5} &
    \textbf{45.5} & \textbf{39.4} & \textbf{31.9}  \\
    
    \palm{}-XXS (Lang) &
    \textbf{7.7} & \textbf{7.6} & \textbf{6.7} &
    37.2 & 26.8 &
    \textbf{17.7} & \textbf{8.8} & \textbf{12.8} &
    \underline{43.6} & \underline{38.3} & \underline{31.5} \\
    
    \midrule
    
    \palm{}-XS (En) &
    0.3 & 0.2 & 0.5 &
    64.3 & \underline{62.2} &
    30.6 & 23.9 & 20.8 &
    35.9 & 32.1 & 27.2 \\
    
    \palm{}-XS (Hi) &
    \underline{3.5} & \underline{5.5} & \underline{9.9} &
    \textbf{65.4} & \textbf{63.5} &
    \underline{33.2} & \underline{25.8} & \underline{22.7} &
    \underline{49.3} & \underline{46.8} & \underline{40.7} \\
    
    \palm{}-XS (Lang) &
    \textbf{18.4} & \textbf{16.4} & \textbf{13.0} &
    \underline{65.1} & 53.3 &
    \textbf{35.8} & \textbf{27.6} & \textbf{26.1} &
    \textbf{53.3} & \textbf{51.5} & \textbf{42.2} \\
    
    \midrule
    
    \palm{}-S (En) &
    0.4 & 0.2 & 0.5 &
    67.4 & \underline{66.8} &
    27.5 & 19.9 & 19.9 &
    48.6 & 47.1 & 40.8 \\
    
    \palm{}-S (Hi) &
    \underline{4.4} & \underline{6.9} & \underline{13.2} &
    \underline{68.5} & \textbf{67.5} &
    \underline{34.2} & \underline{27.0} & \underline{24.9} &
    \underline{58.3} & \underline{57.0} & \underline{49.0} \\
    
    \palm{}-S (Lang) &
    \textbf{22.4} & \textbf{19.8} & \textbf{15.1} &
    \textbf{69.9} & 57.3 &
    \textbf{36.6} & \textbf{30.3} & \textbf{28.6} &
    \textbf{60.1} & \textbf{61.4} & \textbf{53.6} \\
    
    \midrule
    
    \palm{}-L (En) &
    0.4 & 0.2 & 0.6 &
    71.7 & \underline{69.8} &
    37.7 & 33.2 & 29.7 &
    28.7 & 27.5 & 26.2 \\
    
    \palm{}-L (Hi) &
    \underline{4.7} & \underline{7.0} & \underline{13.8} &
    \textbf{72.6} & \textbf{71.0} &
    \underline{39.7} & \underline{34.6} & \underline{31.2} &
    \underline{45.5} & \underline{44.8} & \underline{41.5} \\
    
    \palm{}-L (Lang) &
    \textbf{25.2} & \textbf{23.1} & \textbf{17.5} &
    \underline{72.5} & 59.8 &
    \textbf{41.9} & \textbf{36.7} & \textbf{34.6} &
    \textbf{57.3} & \textbf{57.9} & \textbf{53.9} \\

    \bottomrule
  \end{tabular}
  \caption{\textbf{
  Effect of in-context exemplar language} (\S\ref{ssec:transfer-learning}): Performance comparison when the one-shot exemplar is provided in English (En) or Hindi (Hi) as opposed to the language of the test instance (Lang). In-context prompting in the test language (Lang) provides the best performance, followed by Hindi (Hi) and then English (En). This follows the same order as relatedness between test and prompting language, highlighting the benefit of prompting in a language more related to the test language (e.g., Hindi compared to English in this case).
  }
  \label{tab:one-shot-results-comparison-hi-vs-en-in-context}
\end{table*}

\subsection{In-context learning on \dataset{}}
\label{ssec:in-context-learning}

In this section we aim to understand the impact of the number of \textit{in-context examples} shown to the LLM during few-shot prompting. 

Since \cssum{} and \xquad{} input passages are long, we are only able to perform 0-and-1-shot prompting. For \xorqaxx{} and \xorqa{} we perform 0-to-3-shot prompting, and for \flores{} we perform 0, 1 and 5-shot prompting.

We show performance for \flores{} and \xorqaxx{} in Table~\ref{tab:few_shot_flores_avg}. Other results are shown in appendix~\ref{app:incontext-learning} due to space limitations. Across model families and sizes we observe that increasing the amount of supervision in terms of the in-context examples improves performance.

\begin{table*}[ht!]
\fontsize{7.8}{10pt}\selectfont
  \centering
  \begin{tabular}{lccc|cc|ccc|ccc}
    \toprule
    \multicolumn{1}{l}{} 
    & \multicolumn{3}{c}{\textbf{\cssum{}}} 
    & \multicolumn{2}{c}{\textbf{\xquad{}}} 
    & \multicolumn{3}{c}{\textbf{\xorqaxx{}}} 
    & \multicolumn{3}{c}{\textbf{\xorqa{}}} \\
    \cmidrule(lr){2-4} \cmidrule(lr){5-6} \cmidrule(lr){7-9} \cmidrule(lr){10-12}
    \multicolumn{1}{l}{\textbf{Model}} & \textbf{Higher} & \textbf{Medium} & \textbf{Low} & \textbf{Higher} & \textbf{Medium} & \textbf{Higher} & \textbf{Medium} & \textbf{Low} & \textbf{Higher} & \textbf{Medium} & \textbf{Low}\\
    \midrule
    \multicolumn{12}{l}{\textit{\mt{}} models -- Fine-Tuned} \\
    \midrule
    \mt{}-B  & 19.5 & 18.9 & 15.1 & 46.2 & 30.9 & 3.8 & 4.0 & 5.5 & 31.7 & 31.4 & 30.8\\
    \mt{}-L  & 20.5 & 19.9 & 15.5 & 54.3 & 38.6 & 11.8 & 11.0 & 10.4 & 56.8 & 53.7 & 45.4 \\
    \mt{}-XL & 22.7 & 21.1 & 15.3 & 57.4 & 40.5 & 20.7 & 13.5 & 15.6 & 58.2 & 56.2 & 46.5 \\
    \mt{}-XXL & 25.9 & 24.2 & 10.4 & \textbf{62.0} & \textbf{44.4} & 28.8 & \textbf{23.6} & \textbf{21.9} & \textbf{70.3} & \textbf{68.9} & \textbf{59.1} \\
    \midrule
    \multicolumn{12}{l}{\textit{\palm{}} models - Fine-Tuned} \\
    \midrule
    \palm{}-XXS & 22.5 & 19.7 & 16.5 & 41.2 & 18.1 &  18.1 & 10.9 & 12.9 & 60.2 & 56.9 & 50.9 \\
    \palm{}-XS  & \cellcolor{green!15} \textbf{28.5} & \cellcolor{green!15} \textbf{25.6} & \cellcolor{green!15} \textbf{18.8} & 40.2 & 16.9 &  \textbf{30.4} & \textbf{23.6} & 19.6 & \cellcolor{green!15} 69.1 & \cellcolor{green!15} 66.6 & \cellcolor{green!15} 56.6 \\
    \midrule
    \multicolumn{12}{l}{\textit{\palm{}} models - Few-shot prompted} \\
    \midrule
    \palm{}-XXS$_{FS}$ & 7.7 & 7.6 & 6.7 & 37.2 & 26.8 &  22.7 & 12.3 & 16.4 & 51.6 & 47.1 & 38.4 \\
    \palm{}-XS$_{FS}$ & 18.4 & 16.4 & 13.0 & \cellcolor{green!15} 65.1 & \cellcolor{green!15} 53.3 & \cellcolor{green!15} 39.2 & \cellcolor{green!15} 32.0 & \cellcolor{green!15} 29.5 & 67.0 & 65.3 & 56.5 \\
    \bottomrule
  \end{tabular}
    
    \caption{(Top) \textbf{Fine-tuning performance} of \mt{} and \palm{} models (\S\ref{ssec:fine-tuning}). \textbf{Bold} represents best numbers among fine-tuned models. \palm{} outperforms \mt{} for longer-form generation task (\cssum{}), whereas \mt{} models do well on short answer-span QA tasks. (Bottom) \textbf{Comparison of in-context learning vs. fine-tuning} on \palm{} models. In \colorbox{green!15}{Green}, we highlight the best \palm{} number (among fine-tuned and few-shot).
      For \cssum{} task requiring longer-form generation, fine-tuning outperforms few-shot prompting.}
  \label{tab:fine_tuning_results}
\end{table*}

\subsection{Transfer from high-resource languages}
\label{ssec:transfer-learning}
For languages with no supervised data, one option to improve performance is utilizing existing supervised data another language as in-context exemplars. In this section we aim to study if the language in which the model is prompted plays a role in performance. 

In Table~\ref{tab:one-shot-results-comparison-hi-vs-en-in-context} we show performance when the model is prompted in English vs. Hindi, a representative higher resourced Indic language. For comparison, we also show performance when the in-context exemplar is in the same language as the test instance. 
We find that Hindi in-context exemplars are much more useful for all models as compared to their English counterparts. Surprisingly, for smaller models, performance with Hindi exemplars comes extremely close to prompting in the test language, even better sometimes.

\subsection{Fine-tuning LLMs on \dataset{} and Comparison with In-Context Learning}
\label{ssec:fine-tuning}
As outlined in Section~\ref{sec:dataset_details}, we also release a small, high-quality training set for all tasks in \dataset{} (except \flores{} which only has dev and test sets). This training data can be used to adapt LLMs to downstream tasks in Indic languages via fine-tuning and other training techniques.

Table~\ref{tab:fine_tuning_results} shows our results of fine-tuning \mt{} and \palm{} models and their comparison with in-context learning using \palm{}. We fine-tune each model on training data from all available languages including English, use the development set for early stopping, and report numbers on the test set.
For question-answering tasks that require generating short spans as answers, we find that older generation \mt{} models significantly outperform smaller \palm{} models in most cases.\footnote{Since the parameter count for \palm{} models is not public, we cannot attribute this performance difference to model sizes.}
On \cssum{} which requires generating a longer summary, we find that \palm{} models are more effective.

For Question-Answering tasks, as the model size increases from \palm{}-XXS to \palm{}-XS, we see that in-context learning yields equal or better performance compared to fine-tuning the model. For example, in \xorqaxx{}, as the model size increases from XXS to XS, we see that the gap between few-shot prompting and fine-tuning significantly increases from 2-4\% (in XXS) to 9-10\% (in XS). In the case of \xquad{}, we see that for the larger \palm{}-XS model, its much better to perform in-context learning as compared to fine-tuning, for both medium and high resource Indic languages. For \xorqa{}, in-context learning reaches the fine-tuning performance as model size increases to \palm{}-XS. For the \cssum{}, the gap between fine-tuning and in-context learning is reducing as model size increases, which reinforces that for even larger model sizes, it might be better to learn in-context.

\subsection{Analyzing Tokenizer across Indic languages}
\label{ssec:tokenization-issues}

\begin{figure}[th!]
  \centering
  \includegraphics[width=\columnwidth]{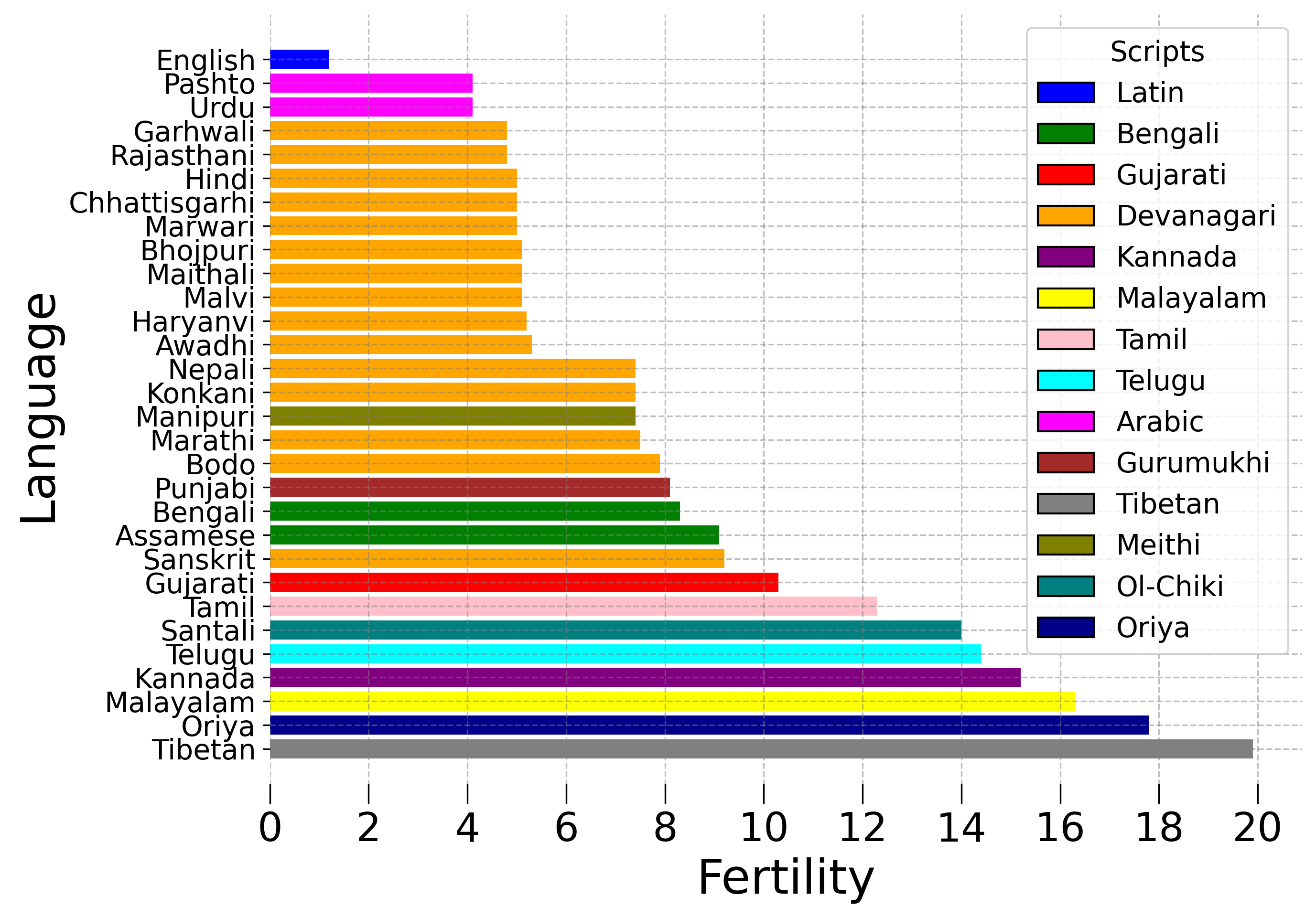}
  \caption{Tokenizer fertility for different languages using OpenAI's Byte Pair Encoding. We note that mid-low resource languages suffer from high token fertility.  (\refsec{ssec:tokenization-issues})}
  \label{fig:tok_fertility}
\end{figure}

\begin{figure}[ht!]
  \centering
  \includegraphics[width=\columnwidth]{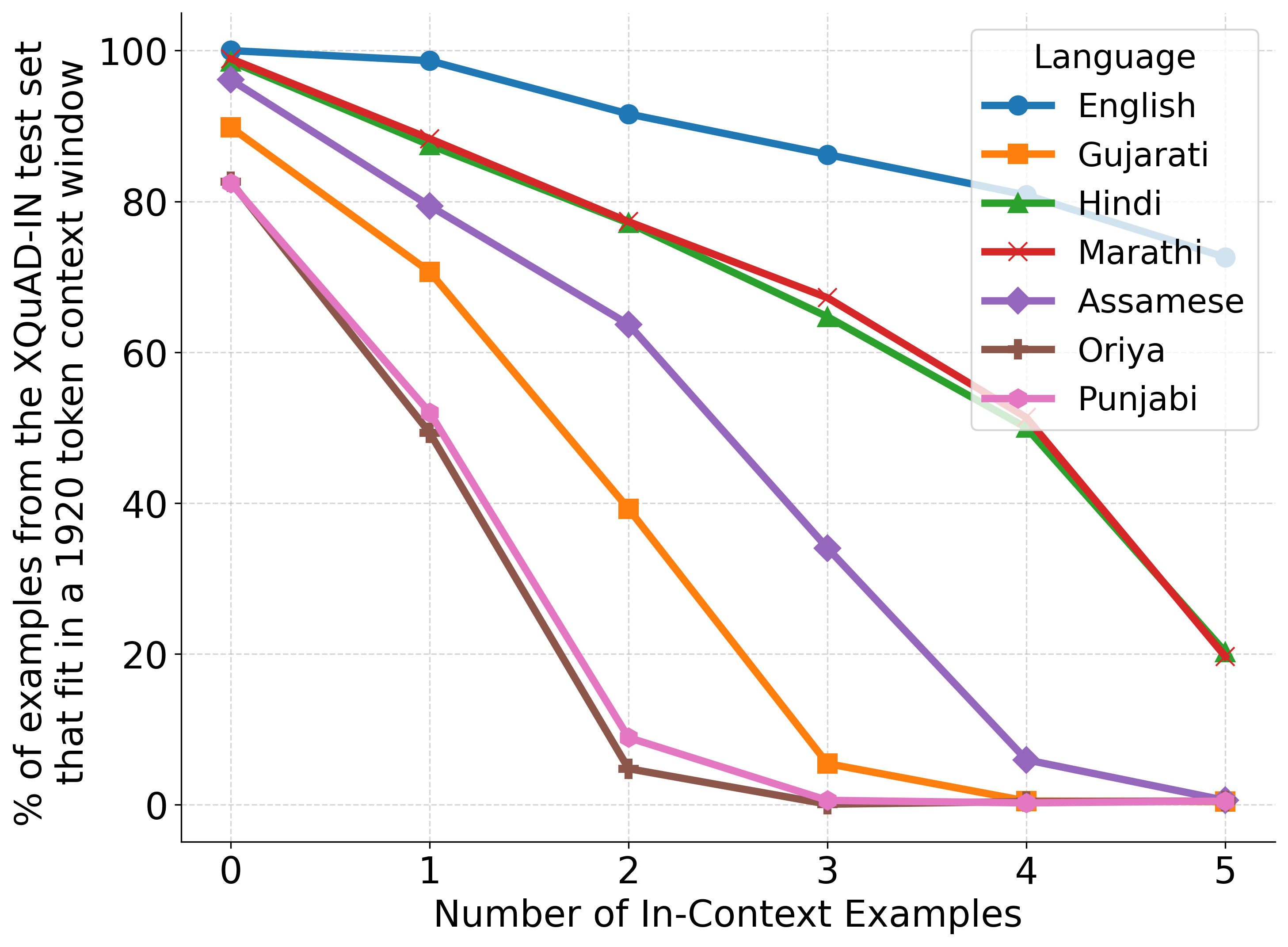}
  \caption{Percentage of the \xquad{} test set in few-shot learning setting that fits in a 1920 token context. High token fertility of mid to low resource languages results in being able to fit much fewer in-context examples compared to higher resourced ones. (\S\ref{ssec:tokenization-issues})
  }
  \label{fig:perc_examples_in_context}
\end{figure}

In Figure \ref{fig:tok_fertility}, we compare the token fertility (average number of sub-words that a word is broken down into by the tokenizer) across all Indic langugaes in \dataset{}.\footnote{We use OpenAI's BPE tokenizer \href{https://platform.openai.com/tokenizer}{(platform.openai.com/tokenizer)}. \palm{} tokenizer is not publicly available.}
We find that the token fertility varies significantly across languages; from 4.1 for Pashto to 19.9 for Tibetan. 

A high token fertility is undesirable and can disproportionately effect a particular language's performance.
For languages where text is broken into more number of tokens, fewer in-context examples can be input to the LLM during inference. This can negatively impact performance (see Table~\ref{tab:few_shot_flores_avg}). In Figure~\ref{fig:perc_examples_in_context}, we show how the percentage of a dataset that fits in a particular context length changes with number of in-context examples for various languages. For example, we see in Figure~\ref{fig:perc_examples_in_context} that for medium resource languages with high token-fertility like Oriya and Punjabi we can in-corporate much fewer in-context examples compared to Indic languages with lower token-fertility like Hindi and Marathi. Analysis in this section is inspired by prior work on studying the impact of tokenization on model utility, inference, and financial cost \citep{Ahia2023DoAL, petrov2023token_unfairness}. Our work supports and compliments these studies for a range of Indic languages. Alternate tokenization and encoding approaches are required to bridge the tokenization disparity between different languages. An example of this is MYTE~\citep{limisiewicz2024myte}, a morphology based byte encoding scheme which leads to more equitable text representations.

\subsection{Qualitative Analysis}
\label{ssec:qualitative-analysis}

We manually analyze predictions from the best performing model \palm{}-L with the aim to understand the shortcomings of current LLMs and highlight areas of improvements for future research. We randomly select 20 examples each in the \cssum{} and \flores{} tasks for the following languages which are reviewed by native speakers: Awadhi, Haryanvi, Chhatisgarhi, Konkani, and Assamese.
We found the following patterns of errors:

\paragraph{Generation in a related language}
The languages Awadhi, Haryanvi, and Chhatisgarhi are related to a higher resource language Hindi and written in the same script Devanagari. We find that the model generates mixed-language output with words mixed from Hindi and also outputs incorrectly inflected forms of the main verbs in the output. 
We show couple of examples of this phenomena in Figure~\ref{fig:qualitative-translate-errors-1} in the appendix.

\paragraph{Hallucination and Missing Information}
In the cross-lingual summarization task \cssum{}, we find that the model often outputs extra information that is not present in the source article. In translation, we have observed examples where come crucial information from the source sentence is missing from the generated output. Also, in some cases, the model fails to understand polysemous English words and generates translation for the incorrect sense.
We show examples of these phenomena in Figures~\ref{fig:qualitative-hallucination-errors-1}, \ref{fig:qualitative-hallucination-errors-2}, and \ref{fig:qualitative-translate-errors-2} in the appendix.
\section{Related Work}
\label{related_work}
In the last few years, many multilingual LLMs have been developed---starting from mBART~\citep{mbart-liu2020multilingual} trained on 25 languages to LLMs that are pre-trained on hundreds of languages, such as \mt{}~\citep{xue-etal-2021-mt5}, \palm{}~\citep{anil2023palm}, \gpt{4}~\citep{achiam2023gpt}, Gemini~\citep{Anil2023GeminiAF}, and others.
These LLMs are typically evaluated on individual multilingual tasks for Translation: WMT~\citep{wmt21}, FLORES~\citep{nllbteam2022language}; Question-Answering: XQuAD~\citep{xquad}, TyDiQA~\citep{tydiqa}, XorQA~\citep{xorqa}; Summarization: XLSUM~\citep{xlsum}; Reasoning: MGSM~\citep{mgsm}, XCOPA \citep{xcopa} to name a few, or on  multilingual benchmarks such as, XTREME~\citep{xtreme} and XTREME-UP~\citep{ruder2023xtreme}.
However, most of these evaluation resources contain only a handful of languages or do not contain data for low resource languages, especially Indics. Besides, cross-lingual evaluation data is even more sparse. This work is an effort to bridge these gaps by releasing \dataset{}, a suite of datasets covering diverse cross-lingual and multilingual generation tasks in Indic languages.

Most work on creating evaluation data on Indic languages  have focused on natural language understanding (NLU) tasks.
\citet{kakwani-etal-2020-indicnlpsuite} and \citet{doddapaneni-etal-2023-towards} have released NLU test sets in Indic languages for a wide variety of tasks such as QA and NLI.
Naamapadam~\citep{mhaske-etal-2023-naamapadam} is a named entity recognition dataset specifically for Indic languages, MASSIVE \citep{fitzgerald2022massive} is a slot-filling and intent classification dataset available in 7 Indic languages, IndicGLUE \citep{kakwani-etal-2020-indicnlpsuite} is an NLU benchmark for 11 Indic languages, whereas GLUECoS \citep{khanuja-etal-2020-gluecos} is a Hindi-English code-mixed benchmark, containing various NLU tasks. The Belebele Benchmark~\citep{bandarkar2023belebele} is a multiple-choice machine reading comprehension dataset for 122 languages of which 17 are Indic.
On the other hand, \dataset{} is a natural language generation (NLG) benchmark.

Recently, there has been work in creating evaluation benchmarks for natural language generation (NLG) on Indic languages.
IndicNLG Suite \citep{kumar-etal-2022-indicnlg}, consisting of 5 NLG taks in 11 Indic languages, is a leap in this direction.
These datasets in this suite are automatically created, either using data from the web (e.g., Wikipedia) or using translation systems. There are few works which create evaluation data for individual tasks in Indic languages. For example,  IndicTrans2~\citep{gala2023indictrans} creates an n-way parallel dataset for machine translation in 22 scheduled Indian Languages, Mukhyansh~\citep{madasu-etal-2023-mukhyansh} and PMIndiaSum~\citep{urlana-etal-2023-pmindiasum} are headline generation datasets for 8 and 14 Indic languages respectively, and TeSum~\citep{urlana-etal-2022-tesum} is an abstractive summarization dataset in the Telugu language. \citet{ramesh-etal-2022-samanantar} introduced Samanantar, a large translation dataset covering 11 Indic languages.
Our work complements IndicNLGSuite and the other datasets in multiple ways.
\dataset{} is manually annotated ensuring high-quality, noise-free text which is not typically found on the web. Our benchmark contains evaluation data for a much larger set of languages spanning low, medium and high resource.
Our datasets are multi-language parallel enabling better comparison among different languages.
Lastly, we focus on a complementary and challenging set of tasks, including cross-lingual summarization, cross-lingual and multilingual question answering, and translation.
\section{Conclusion}
We release \dataset{}, the largest benchmark for evaluating LLMs on 5 user-facing generation tasks across 29 Indic languages, providing evaluation data for many under-represented Indic languages for the first time. \dataset{} is broad coverage along many dimensions -- it covers 13 writing scripts, 4 language families, and spans languages across the available web resource  spectrum. We carry out extensive comparison of current SoTA LLMs on \dataset{} and highlight areas for future improvement. We are hopeful \dataset{} will play an important role in further development of LLMs in Indic languages ultimately benefiting a billion-plus population.
\section{Limitations}
Since \dataset{} extends existing benchmarks to new Indic languages through human translation, it may miss some  India-specific entities and linguistic nuances. Future work can explore trans-localization for creating improved evaluation and fine-tuning. \dataset{} doesn't cover long-form generation and reasoning tasks. Creating such datasets is part of our future work. 
\section*{Acknowledgments}
We thank Aditi Chaudhury, Ashok Popat, Shachi Dave, Sagar Gubbi, Megh Umekar and members of the Languages team at Google Research India (GRI) for providing feedback on this work. The authors would like to thank Manish Gupta and Divy Thakkar for their support and guidance.

\bibliography{custom}

\appendix
\clearpage
\textbf{\LARGE Appendix}

\appendix

\begin{table*}[h!]
    \centering
    \rowcolors{2}{highcolor}{white}
    \begin{tabular}{lllll}
        \toprule
        Code & Language & Script & Family & Resource \\
        \midrule
        en & English & Latin & Germanic & high \\
        \midrule
        \rowcolor{highcolor}
        bn & Bengali & Bengali & Indo-European & higher (Indic relative) \\
        \rowcolor{highcolor}
        gu & Gujarati & Gujarati & Indo-European & higher (Indic relative) \\
        \rowcolor{highcolor}
        hi & Hindi & Devanagari & Indo-European & higher (Indic relative) \\
        \rowcolor{highcolor}
        kn & Kannada & Kannada & Dravidian & higher (Indic relative) \\
        \rowcolor{highcolor}
        ml & Malayalam & Malayalam & Dravidian & higher (Indic relative) \\
        \rowcolor{highcolor}
        mr & Marathi & Devanagari & Indo-European & higher (Indic relative) \\
        \rowcolor{highcolor}
        ta & Tamil & Tamil & Dravidian & higher (Indic relative) \\
        \rowcolor{highcolor}
        te & Telugu & Telugu & Dravidian & higher (Indic relative) \\
        \rowcolor{highcolor}
        ur & Urdu & Arabic & Indo-European & higher (Indic relative) \\
        \rowcolor{mediumcolor}
        as & Assamese & Bengali & Indo-European & med \\
        \rowcolor{mediumcolor}
        bho & Bhojpuri & Devanagari & Indo-European & med \\
        \rowcolor{mediumcolor}
        ne & Nepali & Devanagari & Indo-European & med \\
        \rowcolor{mediumcolor}
        or & Odia & Odia & Indo-European & med \\
        \rowcolor{mediumcolor}
        pa & Punjabi & Gurumukhi & Indo-European & med \\
        \rowcolor{mediumcolor}
        ps & Pashto & Arabic & Indo-European & med \\
        \rowcolor{mediumcolor}
        sa & Sanskrit & Devanagari & Indo-European & med \\
        \rowcolor{lowcolor}
        awa & Awadhi & Devanagari & Indo-European & low \\
        \rowcolor{lowcolor}
        bgc & Haryanvi & Devanagari & Indo-European & low \\
        \rowcolor{lowcolor}
        bo & Tibetan & Tibetan & Sino-Tibetan & low \\
        \rowcolor{lowcolor}
        brx & Bodo & Devanagari & Sino-Tibetan & low \\
        \rowcolor{lowcolor}
        gbm & Garhwali & Devanagari & Indo-European & low \\
        \rowcolor{lowcolor}
        gom & Konkani & Devanagari & Indo-European & low \\
        \rowcolor{lowcolor}
        hne & Chhattisgarhi & Devanagari & Indo-European & low \\
        \rowcolor{lowcolor}
        hoj & Rajasthani & Devanagari & Indo-European & low \\
        \rowcolor{lowcolor}
        mai & Maithili & Devanagari & Indo-European & low \\
        \rowcolor{lowcolor}
        mni & Manipuri & Meithi & Sino-Tibetan & low \\
        \rowcolor{lowcolor}
        mup & Malvi & Devanagari & Indo-European & low \\
        \rowcolor{lowcolor}
        mwr & Marwari & Devanagari & Indo-European & low \\
        \rowcolor{lowcolor}
        sat & Santali & Ol Chiki & Austroasiatic & low \\

        \bottomrule
    \end{tabular}
    \caption{Resource based language classification into relatively higher, medium and low resource for the languages studied in our work. As mentioned previously, we note that the languages classified   higher,  e.g., Hindi or Bengali, are in fact mid-low Web resource when compared to English and other truly high resource languages globally. For example, using Wikipedia as a proxy for language resources, compared to 6.6M+ Wikipedia articles in English, there are only 160K Hindi Wikipedia articles. See Appendix \S\ref{app:lang_resource_classification} for more details.}
    \label{tab:lang_resource_based_classification}
\end{table*}


\section{Resource-wise Language Classification}
\label{app:lang_resource_classification}
See Table~\ref{tab:lang_resource_based_classification} for the language resource based classification of the languages studied in our work. This classification is done on the basis of multiple criteria, such as whether google translate supports the particular language, the classification of Indic languages suggested by \citet{doddapaneni-etal-2023-towards} which in turn uses \citet{joshi-etal-2020-state}.

\section{Detailed and Additonal Experimental Results}
In this section, we report 
(a) one-shot results across all models and tasks averaged over different language categories (\ref{app:lang-resources-allmodels});
(b) performance for different tasks and models by varying the number of in-context exemplars shown in few-shot prompting (\ref{app:few-shot-complete});
(c) one-shot performance for the largest model in each LLM family across all languages (\ref{app:langwise-one-shot}); 
(d) fine-tuning performance for the largest models across all languages (\ref{app:langwise-finetune}).

\subsection{Resource wise one-shot results}
\label{app:lang-resources-allmodels}
In Table~\ref{tab:one-shot-results-lang-resource-allmodels} we show performance for all LLMs considered in this paper on the one-shot prompting method across different language categories.

\subsection{Performance by varying number of in-context examples}
\label{app:few-shot-complete}
In Tables~\ref{tab:few_shot_cssum_avg},~\ref{tab:few_shot_xquad_avg},~\ref{tab:few_shot_xorqa_avg}, and \ref{tab:few_shot_flores_full_avg} we show additional results on how performance changes with the number of in-context examples across various tasks in \dataset{}.

\subsection{Language wise one-shot performance across models}
\label{app:langwise-one-shot}
In Tables~\ref{tab:1shot_cssum_detailed_results}, \ref{tab:1shot_flores_detailed_results}, \ref{tab:1shot_xorqa_detailed_results}, \ref{tab:1shot_xquad_detailed_results} we show language wise breakdown of performance for largest/best models -- \llama{}-65B, \gpt{4}, \palm{}-L in one-shot setting, across all \dataset{} tasks.

\subsection{Language wise fine-tuning performance across models}
\label{app:langwise-finetune}
In Tables~\ref{tab:ft_xorqa_detailed_results}, \ref{tab:ft_cssum_detailed_results}, \ref{tab:ft_xquad_detailed_results} we show language wise breakdown of performance for largest \mt{} and \palm{} models that we fine-tune on \cssum{}, \xquad{}, \xorqaxx{}, \xorqa{}.

\subsection{Indic Specific LLMs}
\label{app:indic-llms}
To the best of our knowledge, there is no LLM that is pre-trained or fine-tuned on a broad set of Indic languages, or a majority of the 29 languages we study in this work. Some Indic LLMs such as OpenHathi (or its instruction-tuned variant \airavata{} \citep{gala2024airavata}), Tamil-LLama \citep{balachandran2023tamilllama}, or Kan-LLama are primarily trained/fine-tuned on a single Indic language (Hindi, Telugu, and Kannada, respectively).

In this section we evaluate an Indic specific LLM, Airavata which is primarily pre-trained and fine-tuned on English and Hindi language data. We evaluate the model on 6 Indic languages: Hindi (hi), Bengali (bn), Punjabi (pa), Assamese (as), Manipuri (mni) and Santali (sat) (2 each from higher, medium and low resource), and English. See Table \ref{tab:one-shot-results-airavata} for the results.

We find that Airavata performs reasonably well only on the Hindi language, and its performance on other Indic languages is much worse (as expected since it is only trained for Hindi). Airavata performs significantly better than LLaMA on the Hindi language, across tasks, except for CSSUM-IN summarization tasks. BLOOMZ outperforms Airavata on most tasks in the Hindi language.

\begin{table*}[h!]
\fontsize{4.7}{10pt}\selectfont
\begin{tabular}{@{}l@{\hspace{0.1cm}}ccc|ccc|ccc|ccc|ccc@{}}
    \toprule
     & \multicolumn{3}{c}{CSSUM-IN} & \multicolumn{3}{c}{FLORES-IN (enxx)} & \multicolumn{3}{c}{XQUAD-IN} & \multicolumn{3}{c}{XORQA-IN-XX} & \multicolumn{3}{c}{XORQA-IN-EN} \\
    \cmidrule(lr){2-4} \cmidrule(lr){5-7} \cmidrule(lr){8-10} \cmidrule(lr){11-13} \cmidrule(lr){14-16}
    Language & \airavata{} & \llama{} & \bloomz{} & \airavata{} & \llama{} & \bloomz{} & \airavata{} & \llama{} & \bloomz{} & \airavata{} & \llama{} & \bloomz{} & \airavata{} & \llama{} & \bloomz{} \\
    \midrule
    en & 31.6 & 19.1 & 13.9  & - & - & - & 75.7 & 49.2 & 86.8  & 69.5 & 56.0 & 72.2  & 69.5 & 59.7 & 68.7  \\
    hi & 0.3  & 6.0  & 1.1   & 55.9 & 19.0 & 66.8  & 48.7 & 16.6 & 64.3  & 46.8 & 30.3 & 18.6  & 31.6 & 19.4 & 67.6  \\
    bn & 0.0  & 2.0  & 0.2   & 8.5  & 9.6  & 73.9  & 6.5  & 2.5  & 57.9  & 0.2  & 3.8  & 0.3   & 33.8 & 12.6 & 70.0  \\
    pa & 0.1  & 1.2  & 0.3   & 6.8  & 6.8  & 55.7  & 3.2  & 0.8  & 60.2  & 2.5  & 0.9  & 4.7   & 21.2 & 1.8  & 67.3  \\
    as & 0.3  & 2.0  & 1.4   & 5.8  & 5.5  & 50.1  & 5.2  & 2.1  & 52.8  & 0.5  & 3.0  & 2.6   & 32.4 & 9.4  & 57.7  \\
    mni  & 0.2  & 2.3  & 0.2   & 6.1  & 6.7  & 6.5   & - & - & - & 1.4  & 2.9  & 0.3   & 22.6 & 5.4  & 21.5  \\
    sat  & 0.0  & 2.4  & 0.3   & 6.4  & 8.3  & 15.9  & - & - & - & 1.1  & 1.1  & 0.0   & 18.6 & 5.7  & 5.5  \\          \bottomrule
    \end{tabular}
  \caption{One-Shot performance on \dataset{} comparing language-specific Indic LLM Airavata with other similar sized open-source models. Airavata is primarily pre-trained and fine-tuned for Hindi and English. As expected, Airavata outperforms other LLMs on Hindi whereas it's performance on other Indic languages is significantly worse than broad-coverage LLMs like BLOOMZ. See Appendix \S\ref{app:indic-llms} for more details.}
  \label{tab:one-shot-results-airavata}
\end{table*}

\begin{table*}[h!]
    \centering
    \begin{tabular}{ll|ccccc}
        \toprule
        Code & Language & \multicolumn{5}{c}{Coverage in \dataset{}} \\
        \cmidrule(lr){3-7}
        & & \small{\cssum{}} & \small{\flores{}} & \small{\xquad{}} & \small{\xorqaxx{}} & \small{\xorqa{}} \\
        \midrule
        \rowcolor{highcolor}
        bn & Bengali & \checkmark & \checkmark & \checkmark & \checkmark & \checkmark \\
        \rowcolor{highcolor}
        gu & Gujarati & \checkmark & \checkmark & \checkmark & \checkmark & \checkmark \\
        \rowcolor{highcolor}
        hi & Hindi & \checkmark & \checkmark & \checkmark & \checkmark & \checkmark \\
        \rowcolor{highcolor}
        kn & Kannada & \checkmark & \checkmark & \checkmark & \checkmark & \checkmark \\
        \rowcolor{highcolor}
        ml & Malayalam & \checkmark & \checkmark & \checkmark & \checkmark & \checkmark \\
        \rowcolor{highcolor}
        mr & Marathi & \checkmark & \checkmark & \checkmark & \checkmark & \checkmark \\
        \rowcolor{highcolor}
        ta & Tamil & \checkmark & \checkmark & \checkmark & \checkmark & \checkmark \\
        \rowcolor{highcolor}
        te & Telugu & \checkmark & \checkmark & \checkmark & \checkmark & \checkmark \\
        \rowcolor{highcolor}
        ur & Urdu & \checkmark & \checkmark & \checkmark & \checkmark & \checkmark \\
        \midrule
        \rowcolor{mediumcolor}
        as & Assamese & \checkmark & \checkmark & \checkmark & \checkmark & \checkmark \\
        \rowcolor{mediumcolor}
        bho & Bhojpuri & \checkmark & \checkmark &  & \checkmark & \checkmark \\
        \rowcolor{mediumcolor}
        ne & Nepali & \checkmark & \checkmark &  &  &  \\
        \rowcolor{mediumcolor}
        or & Odia & \checkmark & \checkmark & \checkmark & \checkmark & \checkmark \\
        \rowcolor{mediumcolor}
        pa & Punjabi & \checkmark & \checkmark & \checkmark & \checkmark & \checkmark \\
        \rowcolor{mediumcolor}
        ps & Pashto & \checkmark & \checkmark &  & \checkmark & \checkmark \\
        \rowcolor{mediumcolor}
        sa & Sanskrit & \checkmark & \checkmark &  & \checkmark & \checkmark \\
        \midrule
        \rowcolor{lowcolor}
        awa & Awadhi & \checkmark & \checkmark &  & \checkmark & \checkmark \\
        \rowcolor{lowcolor}
        bgc & Haryanvi & \checkmark & \checkmark &  & \checkmark & \checkmark \\
        \rowcolor{lowcolor}
        bo & Tibetan & \checkmark & \checkmark &  & \checkmark & \checkmark \\
        \rowcolor{lowcolor}
        brx & Bodo & \checkmark & \checkmark &  & \checkmark & \checkmark \\
        \rowcolor{lowcolor}
        gbm & Garhwali & \checkmark & \checkmark &  & \checkmark & \checkmark \\
        \rowcolor{lowcolor}
        gom & Konkani & \checkmark & \checkmark &  & \checkmark & \checkmark \\
        \rowcolor{lowcolor}
        hne & Chhattisgarhi & \checkmark & \checkmark &  & \checkmark & \checkmark \\
        \rowcolor{lowcolor}
        hoj & Rajasthani & \checkmark & \checkmark &  & \checkmark & \checkmark \\
        \rowcolor{lowcolor}
        mai & Maithili & \checkmark & \checkmark &  & \checkmark & \checkmark \\
        \rowcolor{lowcolor}
        mni & Manipuri & \checkmark & \checkmark &  & \checkmark & \checkmark \\
        \rowcolor{lowcolor}
        mup & Malvi & \checkmark & \checkmark &  & \checkmark & \checkmark \\
        \rowcolor{lowcolor}
        mwr & Marwari & \checkmark & \checkmark &  & \checkmark & \checkmark \\
        \rowcolor{lowcolor}
        sat & Santali & \checkmark & \checkmark &  & \checkmark & \checkmark \\
        \bottomrule
    \end{tabular}
    \caption{Coverage of languages across different tasks in \dataset{}.}
    \label{tab:lang_coverage}
\end{table*}

\begin{table*}[h!]
\fontsize{5.7}{10pt}\selectfont
  \centering
  \begin{tabular}{lccc|cc|ccc|ccc|ccc}
    \toprule
    \multicolumn{1}{l}{} & \multicolumn{3}{c}{\textbf{\xorqa{}}} & \multicolumn{2}{c}{\textbf{\xquad{}}} & \multicolumn{3}{c}{\textbf{\cssum{}}} & \multicolumn{3}{c}{\textbf{\xorqaxx{}}} & \multicolumn{3}{c}{\textbf{\flores{}}}\\
    \cmidrule(lr){2-4} \cmidrule(lr){5-6} \cmidrule(lr){7-9} \cmidrule(lr){10-12} \cmidrule(lr){13-15}
    \multicolumn{1}{l}{\textbf{Model}} & \textbf{High} & \textbf{Medium} & \textbf{Low} & \textbf{High} & \textbf{Medium} & \textbf{High} & \textbf{Medium} & \textbf{Low} & \textbf{High} & \textbf{Medium} & \textbf{Low} & \textbf{High} & \textbf{Medium} & \textbf{Low}\\
    
    \midrule
    \llama{}-7B & 9.9 & 7.6 & 12.2 & 4.7 & 1.1 & 3.6 & 3.6 & 3.7 & 8.1 & 3.4 & 8.8 & 11.2/20.4 & 9.6/20.2 & 12.8/23.2 \\
    \llama{}-13B & 10.5 & 9.9 & 14.2 & 5.5 & 1.5 & 3.9 & 4.1 & 4.2 & 9.6 & 6.3 & 12.8 & 12.6/22.5 & 11.3/22.0 & 14.8/26.4 \\
    \llama{}-65B & 16.4 & 14.0 & 17.3 & 8.8 & 1.9 & 4.4 & 4.6 & 4.7 & 17.7 & 13.5 & 17.1 & 18.2/31.5 & 15.4/30.0 & 19.5/35.0 \\
    
    \midrule
    \gemma{}-7B-PT & 25.7 & 22.5 & 23.1 & 0.6 & 0.1 & 0.0 & 0.0 & 0.0 & 16.5 & 6.9 & 10.5 & 40.7 / 58.8 & 25.5 / 49.5 & 29.8 / 45.0 \\
    \gemma{}-7B-IT & 29.5 & 23.9 & 21.9 & 38.8 & 24.8 & 13.9 & 11.5 & 10.0 & 18.9 & 8.3 & 12.2 & 17.6 / 33.7 & 15.0 / 26.1 & 21.3 / 27.7 \\

    \midrule    
    \bloom{}-7B & 29.3 & 21.6 & 20.6 & 15.2 & 9.8 & 3.1 & 3.8 & 4.3 & 10.8 & 5.0 & 7.2 & 19.8 / 34.7 & 16.3 / 31.4 & 18.4 / 28.6 \\
    \bloomz{}-7B & \textbf{64.7} & 45.8 & 39.5 & 55.5 & 48.1 & 1.5 & 1.7 & 0.6 & 10.8 & 2.8 & 6.2 & \textbf{67.7} / 59.1 & 39.4 / 50.2 & 22.9 / 40.0 \\
    
    \midrule
    GPT-3.5 & 39.0 & 34.9 & 33.4 & 36.2 & 23.9 & 17.6 & 16.5 & 15.3 & 24.2 & 20.0 & 20.6 & 32.9/52.9 & 27.3/48.2 & 27.6/43.8 \\
    GPT-4 & 49.4 & 50.0 & 41.8 & 56.1 & 54.6 & 19.4 & 17.9 & 16.3 & 25.8 & 21.6 & 22.6 & 36.2/59.6 & 30.7/55.2 & 29.9/50.5 \\
    
    \midrule 
    \palm{}-XXS & 43.6 & 38.3 & 31.5 & 37.2 & 26.8 & 7.7 & 7.6 & 6.7 & 17.7 & 8.8 & 12.8 & 31.7 / 52.1 & 19.0 / 43.2 & 21.4 / 37.6 \\
    \palm{}-XS & 53.3 & 51.5 & 42.2 & 65.1 & 53.3 & 18.4 & 16.4 & 13.0 & 35.8 & 27.6 & 26.1 & 52.0 / 64.8 & 39.0 / 60.6 & 33.7 / 52.6 \\
    \palm{}-S & 60.1 & \textbf{61.4} & 53.6 & 69.9 & 57.3 & 22.4 & 19.8 & 15.1 & 36.6 & 30.3 & 28.6 & 54.7 / 66.8 & 42.5 / 63.5 & 36.4 / 57.0 \\
    \palm{}-L & 57.3 & 57.9 & \textbf{53.9} & \textbf{72.5} & \textbf{59.8} & \textbf{25.2} & \textbf{23.1} & \textbf{17.5} & \textbf{41.9} & \textbf{36.7} & \textbf{34.6} & 56.9 / \textbf{68.2} & \textbf{45.9 / 65.6} & \textbf{41.9 / 62.6} \\
    \bottomrule
  \end{tabular}
  \caption{\textbf{One-Shot performance on \dataset{}}. Results are averaged over high, medium and low resource language examples present in the dataset. For question answering we measure the F1/EM score while for summarization and translation we measure ChrF score. See Appendix \S\ref{app:lang-resources-allmodels} for more details.}
  \label{tab:one-shot-results-lang-resource-allmodels}
\end{table*}

\label{app:incontext-learning}
\begin{table}[ht!]
\fontsize{9}{11pt}\selectfont
  \centering
  \begin{tabular}{lcc}
    \toprule
    & \multicolumn{2}{c}{\textbf{Shots}} \\
    \cmidrule(lr){2-3}
    \textbf{Model (LLM)} & \textbf{0} & \textbf{1} \\
    \midrule
    \llama{}-7B & 2.4 & 3.7\\
    \llama{}-13B & 1.5 & 4.1\\
    \llama{}-65B & 5.1 & 4.6\\
    \midrule
    \palm{}-XXS & 0.2 & 7.2\\
    \palm{}-XS & 1.0 & 15.5\\
    \palm{}-S & 4.1 & 18.5\\
    \palm{}-L & \textbf{7.7} & \textbf{21.2} \\
    \bottomrule
  \end{tabular}
  \caption{Few-shot results for \palm{} models on \cssum{} dataset. ChrF scores are reported.}
  \label{tab:few_shot_cssum_avg}
\end{table}

\begin{table}[ht!]
\fontsize{9}{11pt}\selectfont
  \centering
  \begin{tabular}{lcc}
    \toprule
    & \multicolumn{2}{c}{\textbf{Shots}} \\
    \cmidrule(lr){2-3}
    \textbf{Model (LLM)} & \textbf{0} & \textbf{1} \\
    \midrule
    \llama{}-7B & 4.3 & 3.8\\
    \llama{}-13B & 4.7 & 4.5\\
    \llama{}-65B & 7.2 & 7.1\\
    \midrule
    \palm{}-XXS & 19.0 & 34.6\\
    \palm{}-XS & 38.7 & 62.2\\
    \palm{}-S & 44.7 & 66.7\\
    \palm{}-L & \textbf{50.6} & \textbf{69.3} \\
    \bottomrule
  \end{tabular}
  \caption{Few-shot results for \palm{} models on \xquad{} dataset. F1 scores are reported.}
  \label{tab:few_shot_xquad_avg}
\end{table}

\begin{table}[ht!]
\fontsize{9}{11pt}\selectfont
  \centering
  \begin{tabular}{lcccc}
    \toprule
    & \multicolumn{4}{c}{\textbf{Shots}} \\
    \cmidrule(lr){2-5}
    \textbf{Model (LLM)} & \textbf{0} & \textbf{1} & \textbf{2} & \textbf{3} \\
    \midrule
    \llama{}-7B & 4.7 & 10.4 & 12.5 & 14.1 \\
    \llama{}-13B & 2.7 & 12.1 & 15.4 & 17.0 \\
    \llama{}-65B & 8.2 & 16.3 & 18.8 & 19.8 \\
    \midrule
    \palm{}-XXS & 31.0 & 36.8 & 41.6 & 44.5 \\
    \palm{}-XS & 15.6 & 47.8 & 57.8 & 61.7 \\
    \palm{}-S & \textbf{40.6} & \textbf{57.4} & 63.4 & 66.3 \\
    \palm{}-L & 31.4 & 55.9 & \textbf{66.3} & \textbf{70.3} \\
    \bottomrule
  \end{tabular}
  \caption{Few-shot results for \palm{} models on \xorqa{} dataset. F1 scores are reported.}
  \label{tab:few_shot_xorqa_avg}
\end{table}

\begin{table}[ht!]
\fontsize{9}{11pt}\selectfont
  \centering
  \begin{tabular}{lccc}
    \toprule
     & \multicolumn{3}{c}{\textbf{\# of in-context examples}} \\
    \cmidrule(lr){2-4}
    \textbf{Model (LLM)} & \textbf{0} & \textbf{1} & \textbf{5} \\
    \midrule
    \llama{}-7B & 8.0 / 10.4 & 11.5 / 21.6 & 11.4 / 19.8 \\
    \llama{}-13B & 8.6 / 20.6 & 13.3 / 24.1 & 13.4 / 22.7 \\
    \llama{}-65B & 14.0 / 28.0 & 18.1 / 32.7 & 18.3 / 30.3 \\
    \midrule
    \palm{}-XXS & 0.8 / 22.4 & 24.0 / 43.4  & 26.9 / 44.8 \\
    \palm{}-XS & 20.1 / 53.3 & 40.7 / 58.3 & 42.3 / 58.8 \\
    \palm{}-S & 24.9 / 60.1 & 43.5 / 61.6 & 45.2 / 61.9 \\
    \palm{}-L & \textbf{31.1} / \textbf{62.2} & \textbf{47.5} / \textbf{65.1} & \textbf{49.3} / \textbf{65.7} \\
    \bottomrule
  \end{tabular}
  \caption{Few-shot results for \llama{} and \palm{} models on \flores{} (both \texttt{enxx} / \texttt{xxen} direction).}
  \label{tab:few_shot_flores_full_avg}
\end{table}

\section{Hyperparameters}
\label{app:hyperparameters}
For Fine-tuning experiments on \mt{}, we hyperparameter search for batch size in the range \{16, 32, 64\} and learning rate in \{5e-3, 1e-3, 5e-4, 1e-4, 5e-5, 1e-5\} and select the best hyperparameters based on the Token-F1 (for Question answering datasets) score or ChrF (for \cssum{}, \flores{} experiments) score on the validation set of the dataset. For \palm{} fine-tuning experiments, we hyperparameter search for batch size in the range \{16, 32, 64\} and learning rate in \{5e-4, 1e-4, 5e-5, 1e-5\} and select the best hyperparameters based on the Token-F1 score or ChrF score, as in the case of \mt{} experiments.
We keep temperature=0 as the default setting while decoding using \palm{} and \gptfam{} family of models. For \llama{} models we increased the temperature to 0.1, since otherwise, the performance of \llama{} were coming out to be close to 0 for many of our datasets.
For each dataset, we fix a prompt for all models since searching for a prompt would have led to a much larger compute cost for training and evaluation. We also choose in-context exemplars randomly from the training set (from the dev set in the case of \flores{}). This decision is a limitation of this work, since it has been shown that prompt and in-context exemplars can significantly impact an LLMs performance \citep{Zhao2021CalibrateBU, zhao-etal-2021-closer}. The prompts we use are provided in \S\ref{app:prompts_used}. Searching for better prompts and exemplars is left as future work.
For all fine-tuning runs, we just perform one run (as opposed to taking average of multiple runs) due to prohibitive costs of fine-tuning LLMs multiple times.

\section{Computing Infra}
\label{app:compute_infra}
For \mt{} fine-tuning runs, we use a combination of 4 to 64 TPU-V3, and for \palm{} we use up to 256 TPU-V4. All experiments take about 6 hours to  1-2 days of time depending upon the resources and size of the model being trained. We use NVIDIA-V100 16GB GPUs for evaluating open-source models like \bloom{}, \bloomz{}, \airavata{} on \dataset{}.

\section{Dataset Licenses}
In compliance with licenses of the original datasets, we release our evaluation datasets under the following licenses:
(a) \xquad{} and \flores{} under the CC BY-SA 4.0 license; (b) \cssum{} under the CC BY-NC-SA 4.0 license; and (d) \xorqain{} under the MIT license.

\section{Prompts}
\label{app:prompts_used}
The set of prompts used for evaluations on all datasets are shown in Table \ref{tab:prompts}.

\begin{figure*}[h!]
     \centering
     \begin{subfigure}{\linewidth}
         \centering
         \fbox{\includegraphics[width=\linewidth]{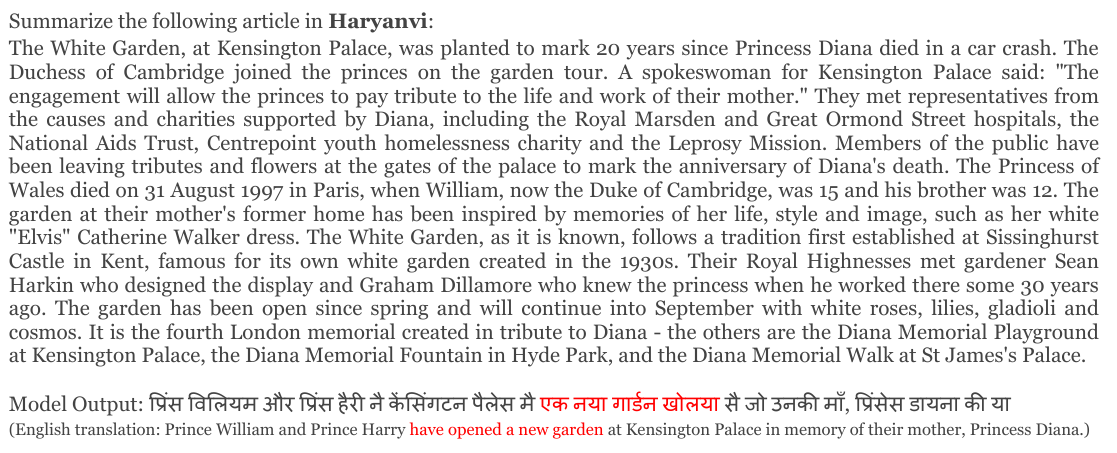}}
         \caption{Model hallucinates that Prince William and Harry opened a new garden. }
         \label{fig:qualitative-hallucination-errors-1}
     \end{subfigure}
     \begin{subfigure}{\linewidth}
         \centering
         \fbox{\includegraphics[width=\linewidth]{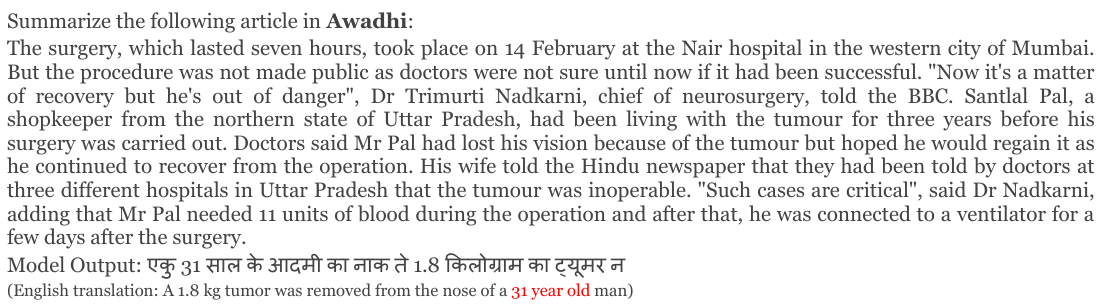}}
         \caption{Model hallucinates the age of the person in the article.}
         \label{fig:qualitative-hallucination-errors-2}
     \end{subfigure}
\caption{Example predictions from \palm{}-L model on the \cssum{} task highlighting issues with hallucinations in model predictions.}
\label{fig:qualitative-hallucination-errors}
\end{figure*}

\begin{figure*}[h!]
     \centering
     \begin{subfigure}{\linewidth}
         \centering
         \fbox{\includegraphics[width=\linewidth]{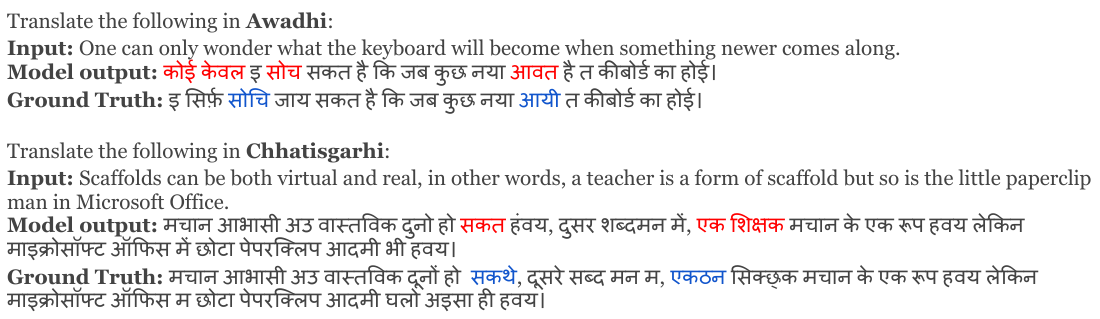}}
         \caption{In these examples of translating into Awadhi and Chhatisgarhi, both low resource languages similar to a higher resource language Hindi, we find that many Hindi words and incorrect inflections for verbs are produced.}
         \label{fig:qualitative-translate-errors-1}
     \end{subfigure}
     \begin{subfigure}{\linewidth}
         \centering
         \fbox{\includegraphics[width=\linewidth]{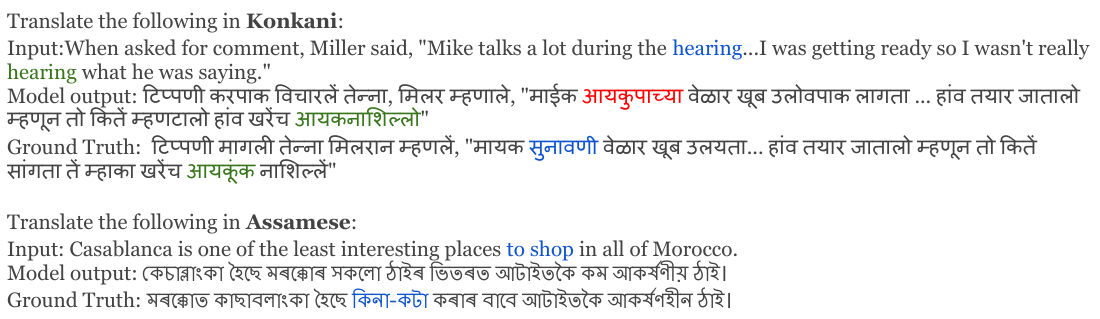}}
         \caption{1. In the first example of translation into Konkani, the model does not understand that the first instance of ``hearing'' refers to a court hearing and the second instance refers to ``listening''. The model incorrectly outputs words related to the ``listening'' sense in both instances. 2. In the translation into Assamese example, the model does not produce translation for the crucial information ``to shop''.}
         \label{fig:qualitative-translate-errors-2}
     \end{subfigure}
\caption{Example predictions from \palm{}-L model on the \flores{} task highlighting issues of (a) producing words in the wrong (higher-resourced) language or words with the wrong inflection, and (b) outputting the incorrect translations for polysemous words in English or missing crucial information from the generated translation.}
\label{fig:qualitative-translate-errors}
\end{figure*}

\begin{table*}[h!]
    \centering
    \begin{tabular}{lp{10cm}}
    \toprule
    \textbf{Dataset} & \textbf{Prompt} \\
    \midrule
    \cssum{} & I will first show a news article in English and then provide a summary of it in the {\color{blue} [Target Language Name]} language.
    \newline
    \newline
    \textbf{Summarize the following article:} {\color{blue} [Article]} \newline \textbf{Summary:}\\
    \midrule
    \flores{}, \texttt{xxen} & Translate the following:
    \newline
    \newline
    \textbf{To English:} {\color{blue} [Sentence in Target Language]} \newline \textbf{Output:}\\
    \flores{}, \texttt{enxx} & Translate the following: 
    \newline\newline
    \textbf{To} {\color{blue} [Target Language Name]:} {\color{blue} [Sentence in English]} 
    \newline 
    \textbf{Output:}\\
    \midrule
    \xquad{} & \vspace{-0.3cm}
    {\color{blue}[Passage in Target Language]} 
    \newline
    \newline 
    \textbf{Q:} {\color{blue} [Question in Target Language]} \newline \textbf{A:}\\
    \midrule
    \xorqaxx{} & Generate an answer in {\color{blue} [Target Language Name]} for the question based on the given passage:
    \newline
    \newline 
    {\color{blue} [Passage in English]}
    \newline
    \newline
    \textbf{Q:} {\color{blue} [Question in Target Language]} \newline \textbf{A:}\\
    \midrule
    \xorqaxx{} & Generate an answer in English for the question based on the given passage:
    \newline
    \newline
    {\color{blue} [Passage in English]} 
    \newline
    \newline
    \textbf{Q:} {\color{blue} [Question in Target Language]} \newline \textbf{A:}\\ \bottomrule
    \end{tabular}
    \caption{Prompt templates used for different datasets in \dataset{}. n-shot examples have the same format as the last, test example given to the model, which comes after the n-shot examples.}
    \label{tab:prompts}
\end{table*}

\newpage
\begin{table*}[h!]
\centering
\begin{tabular}{lccccc}
\toprule
Lang. Code & \multicolumn{5}{c}{\textbf{\cssum{}}} \\
\cmidrule(lr){2-6}
& \textbf{\llama{}-65B} & \textbf{\gemma{}-7B-IT} & \textbf{\bloomz{}-7B} & \textbf{\gpt{4}} & \textbf{\palm{}-L} \\
\midrule
en & 26.1 & 24.8 & 18.6 & 30.3 & 41.1 \\
bn & 2.7 & 13.6 & 0.4 & 20.9 & 23.4 \\
gu & 2.1 & 10.3 & 1.0 & 16.5 & 19.2 \\
hi & 7.8 & 16.8 & 0.9 & 23.1 & 30.4 \\
kn & 2.9 & 13.6 & 0.9 & 15.1 & 25.4 \\
ml & 4.3 & 11.4 & 0.5 & 14.4 & 23.3 \\
mr & 4.1 & 13.6 & 0.6 & 22.4 & 25.4 \\
ta & 5.4 & 20.7 & 2.1 & 20.4 & 28.9 \\
te & 2.5 & 13.6 & 0.7 & 16.6 & 21.9 \\
ur & 7.5 & 13.0 & 6.4 & 24.9 & 28.9 \\
as & 2.7 & 12.5 & 1.8 & 17.3 & 24.3 \\
bho & 5.5 & 13.2 & 0.9 & 19.3 & 20.2 \\
ne & 7.4 & 16.8 & 2.0 & 23.2 & 29.1 \\
or & 2.5 & 4.3 & 0.3 & 10.7 & 23.4 \\
pa & 1.7 & 10.7 & 0.2 & 15.8 & 19.1 \\
ps & 7.0 & 9.2 & 6.3 & 19.9 & 25.2 \\
sa & 5.3 & 13.5 & 0.5 & 19.4 & 20.1 \\
awa & 5.1 & 12.8 & 0.6 & 18.2 & 19.4 \\
bgc & 5.3 & 12.5 & 0.7 & 17.9 & 20.2 \\
bo & 1.5 & 2.8 & 0.1 & 8.5 & 10.0 \\
brx & 4.9 & 2.8 & 0.4 & 15.5 & 11.0 \\
gbm & 5.8 & 11.4 & 2.0 & 16.2 & 16.2 \\
gom & 5.8 & 12.6 & 0.8 & 20.3 & 23.1 \\
hne & 5.5 & 13.4 & 0.9 & 19.2 & 22.0 \\
hoj & 5.2 & 12.5 & 0.3 & 17.7 & 17.7 \\
mai & 4.9 & 12.6 & 0.5 & 18.2 & 21.1 \\
mni & 3.3 & 4.4 & 0.2 & 13.2 & 15.6 \\
mup & 5.3 & 13.1 & 0.6 & 20.0 & 23.8 \\
mwr & 5.1 & 13.2 & 0.8 & 18.4 & 20.2 \\
sat & 2.7 & 5.5 & 0.2 & 8.5 & 7.4 \\
\midrule
\textbf{Avg.} & 4.6 & 11.6 & 1.2 & 17.6 & 21.2 \\
\textbf{High Avg.} & 4.4 & 13.9 & 1.5 & 19.4 & 25.2 \\
\textbf{Medium Avg.} & 4.6 & 11.5 & 1.7 & 17.9 & 23.1 \\
\textbf{Low Avg.} & 4.7 & 10.0 & 0.6 & 16.3 & 17.5 \\
\bottomrule
\end{tabular}
\caption{Performance comparison of models on \cssum{} in one-shot setting across all supported languages.}
\label{tab:1shot_cssum_detailed_results}
\end{table*}

\begin{table*}[h!]
\fontsize{6}{10pt}\selectfont
\centering
\begin{tabular}{l@{}cccccccccc}
\toprule
Lang. Code & \multicolumn{5}{c}{\textbf{\flores{} \texttt{(enxx)}}} & \multicolumn{5}{c}{\textbf{\flores{} \texttt{(xxen)}}} \\
\cmidrule(lr){2-6} \cmidrule(lr){7-11}
 & \textbf{\llama{}-65B} & \textbf{\gemma{}-7B-IT} & \textbf{\bloomz{}-7B} & \textbf{\gpt{4}} & \textbf{\palm{}-L}  & \textbf{\llama{}-65B} & \textbf{\gemma{}-7B-IT} & \textbf{\bloomz{}-7B} & \textbf{\gpt{4}} & \textbf{\palm{}-L}  \\
\midrule
bn  &	18.7 &	27.3 & 71.9 &  40.0 &	54.0 &	37.4 &	35.1 & 60.4 & 59.9 &	 66.8 \\
gu  &	11.3 &	3.3 & 68.1 &  31.3 &	56.1 &	22.1 &	29.1 & 60.9 & 61.2 &	 70.9 \\
hi  &	33.2 &	35.7 & 65.2 &  48.6 &	60.2 &	53.7 &	48.4 & 57.3 & 65.2 &	 70.5 \\
kn  &	11.1 &	15.6 & 66.8 &  29.9 &	58.0 &	20.6 &	27.6 & 59.1 & 57.6 &	 65.7 \\
ml  &	14.5 &	7.1 & 66.2 &  28.4 &	59.7 &	22.8 &	28.8 & 60.6 & 58.6 &	 68.0 \\
mr  &	22.7 &	22.3 & 68.6 &  39.2 &	53.1 &	38.2 &	31.7 & 58.8 & 59.8 &	 68.5 \\
ta  &	16.8 &	29.8 & 75.2 &  34.5 &	60.0 &	24.5 &	30.5 & 55.7 & 53.9 &	 65.7 \\
te  &	10.7 &	16.1 & 68.8 &  30.7 &	60.5 &	21.7 &	33.0 & 56.1 & 58.1 &	 70.5 \\
ur  &	24.6 &	1.3 & 58.5 &  43.5 &	50.6 &	42.8 &	38.8 & 62.8 & 62.0 &	 67.2 \\
awa  &	26.4 &	29.1 & 32.1 &  38.8 &	49.9 &	46.1 &	36.9 & 56.3 & 61.8 &	 69.3 \\
bgc  &	26.6 &	28.9 & 31.1 &  38.0 &	49.8 &	45.1 &	33.3 & 55.7 & 63.5 &	 72.1 \\
bo  &	6.6 &	4.5 & 13.7 &  17.0 &	40.9 &	16.9 &	17.0 & 16.1 & 27.7 &	 50.1 \\
brx  &	12.4 &	13.0 & 12.1 &  17.2 &	11.3 &	20.7 &	17.8 & 17.0 & 29.2 &	 31.0 \\
gbm  &	24.0 &	26.6 & 28.3 &  36.2 &	47.9 &	43.0 &	31.2 & 49.6 & 62.2 &	 72.1 \\
gom  &	17.8 &	19.0 & 17.3 &  27.5 &	39.0 &	29.7 &	23.3 & 28.3 & 48.2 &	 63.2 \\
hne  &	25.4 &	29.2 & 29.3 &  38.4 &	52.7 &	42.8 &	32.9 & 53.6 & 62.3 &	 75.1 \\
hoj  &	23.7 &	26.6 & 28.1 &  36.0 &	48.1 &	41.9 &	30.4 & 51.3 & 62.3 &	 73.0 \\
mai  &	21.4 &	26.0 & 22.8 &  36.2 &	54.1 &	41.8 &	32.7 & 57.5 & 60.9 &	 73.3 \\
mni  &	8.3 &	9.3 & 8.7 &  15.2 &	19.5 &	20.7 &	18.1 & 18.7 & 31.8 &	 41.7 \\
mup  &	27.0 &	28.3 & 31.7 &  39.9 &	51.9 &	45.1 &	34.9 & 49.2 & 62.3 &	 73.0 \\
mwr  &	26.4 &	29.7 & 33.9 &  39.2 &	52.2 &	44.6 &	34.6 & 51.3 & 64.2 &	 74.5 \\
sat  &	8.1 &	6.5 & 8.4 &  9.4 &	27.9 &	16.4 &	17.0 & 16.1 & 20.6 &	 45.6 \\
as  &	9.5 &	18.1 & 51.1 &  28.5 &	44.1 &	25.9 &	23.6 & 60.0 & 51.7 &	 63.3 \\
bho  &	23.4 &	25.1 & 24.9 &  33.5 &	42.9 &	40.9 &	32.0 & 51.8 & 55.3 &	 61.7 \\
ne  &	25.9 &	24.7 & 72.0 &  45.8 &	57.8 &	42.3 &	34.6 & 63.3 & 62.6 &	 72.3 \\
or  &	9.4 &	5.2 & 45.7 &  19.6 &	52.9 &	19.6 &	17.1 & 63.4 & 56.9 &	 68.0 \\
pa  &	9.7 &	2.6 & 57.1 &  30.7 &	51.8 &	22.1 &	28.4 & 60.8 & 62.5 &	 70.1 \\
ps  &	13.3 &	10.9 & 9.2 &  27.3 &	37.6 &	27.7 &	22.0 & 19.4 & 45.7 &	 64.5 \\
sa  &	16.9 &	18.6 & 15.7 &  29.3 &	34.3 &	31.4 &	25.1 & 32.5 & 51.5 &	 59.4 \\
\midrule
\textbf{Avg.} &	18.1 &	18.6 & 40.8 &  32.1  &	47.5 &	32.7 &	29.2 & 48.4 & 54.5 &	 65.1 \\
\textbf{High Avg.} &	18.2 &	17.6 & 67.7 &  36.2  &	56.9 &	31.5 &	33.7 & 59.1 & 59.6 & 68.2 \\
\textbf{Medium Avg.} &	15.4 &	15.0 & 39.4 &  30.7  &	45.9 &	30.0 &	26.1 & 50.2 & 55.2 & 65.6 \\
\textbf{Low Avg.} &	19.5 &	21.3 & 22.9 &  29.9  &	41.9 &	35.0 &	27.7 & 40.0 & 50.5 & 62.6 \\
\bottomrule
\end{tabular}
\caption{Performance comparison of models on \flores{} in one-shot setting across all supported languages.}
\label{tab:1shot_flores_detailed_results}
\end{table*}

\begin{table*}[h!]
\fontsize{6}{10pt}\selectfont

\centering
\begin{tabular}{l@{}cccccccccc}
\toprule
Lang. Code & \multicolumn{5}{c}{\textbf{\xorqaxx{}}} & \multicolumn{5}{c}{\textbf{\xorqa{}}} \\
\cmidrule(lr){2-6} \cmidrule(lr){7-11}
 & \textbf{\llama{}-65B} & \textbf{\gemma{}-7B-IT} & \textbf{\bloomz{}-7B} & \textbf{\gpt{4}} & \textbf{\palm{}-L}  & \textbf{\llama{}-65B} & \textbf{\gemma{}-7B-IT} & \textbf{\bloomz{}-7B} & \textbf{\gpt{4}} & \textbf{\palm{}-L}  \\
\midrule
en & 61.1 & 35.5 & 68.7 & 37.4 & 71.4 & 63.9 & 34.7 & 69.7 & 37.9 & 71.5 \\
bn & 15.8 & 9.4 & 0.6 & 16.5 & 40.4 & 16.9 & 25.7 & 68.4 & 46.2 & 55.6 \\
gu & 8.8 & 18.3 & 11.2 & 23.8 & 40.2 & 12.9 & 32.7 & 64.2 & 51.9 & 53.7 \\
hi & 43.5 & 29.6 & 16.5 & 38.3 & 56.5 & 23.1 & 31.1 & 66.1 & 41.6 & 58.8 \\
kn & 8.7 & 19.4 & 8.8 & 28.8 & 41.1 & 9.4 & 26.2 & 63.5 & 54.6 & 59.6 \\
ml & 13.9 & 26.7 & 19.6 & 31.1 & 56.7 & 16.9 & 30.7 & 66.4 & 55.3 & 54.8 \\
mr & 28.1 & 26.4 & 17.7 & 30.4 & 44.5 & 22.3 & 29.1 & 64.0 & 47.9 & 57.4 \\
ta & 16.0 & 17.5 & 10.7 & 25.4 & 40.2 & 19.0 & 26.8 & 64.0 & 50.1 & 57.7 \\
te & 4.0 & 15.3 & 11.4 & 18.5 & 25.6 & 6.0 & 29.3 & 67.6 & 52.0 & 56.3 \\
ur & 20.6 & 7.6 & 1.1 & 18.9 & 31.6 & 21.5 & 34.1 & 58.5 & 44.9 & 61.5 \\
as & 17.6 & 10.8 & 2.0 & 26.4 & 45.5 & 12.5 & 22.3 & 60.0 & 51.6 & 59.3 \\
bho & 19.9 & 9.9 & 2.0 & 23.8 & 32.1 & 21.4 & 26.7 & 55.2 & 46.9 & 59.4 \\
or & 4.0 & 1.6 & 6.2 & 22.1 & 35.9 & 7.7 & 14.8 & 48.0 & 54.6 & 52.6 \\
pa & 6.9 & 13.3 & 4.0 & 25.7 & 40.9 & 10.7 & 30.4 & 63.2 & 47.3 & 55.5 \\
ps & 10.8 & 3.4 & 0.5 & 12.4 & 25.7 & 12.4 & 23.9 & 3.5 & 44.7 & 62.9 \\
sa & 21.9 & 11.0 & 2.1 & 19.0 & 40.3 & 19.0 & 25.5 & 44.8 & 54.8 & 57.8 \\
awa & 26.3 & 13.1 & 7.8 & 28.5 & 41.3 & 23.3 & 25.3 & 58.1 & 50.3 & 58.4 \\
bgc & 18.5 & 16.4 & 10.2 & 20.1 & 28.1 & 23.1 & 25.6 & 55.7 & 43.8 & 58.4 \\
bo & 3.3 & 0.9 & 5.4 & 32.5 & 32.5 & 3.2 & 14.4 & 6.0 & 33.8 & 52.6 \\
brx & 5.3 & 12.7 & 0.9 & 15.5 & 20.0 & 12.1 & 15.6 & 7.6 & 35.9 & 39.1 \\
gbm & 13.7 & 14.9 & 9.5 & 18.5 & 35.4 & 21.6 & 24.9 & 50.7 & 46.1 & 60.2 \\
gom & 15.1 & 8.0 & 0.8 & 19.2 & 33.2 & 14.0 & 22.1 & 37.8 & 41.4 & 54.2 \\
hne & 29.7 & 22.7 & 12.6 & 32.0 & 43.4 & 22.5 & 24.9 & 56.4 & 43.9 & 54.8 \\
hoj & 31.2 & 25.3 & 17.3 & 36.3 & 53.8 & 21.5 & 23.9 & 51.5 & 48.8 & 62.6 \\
mai & 28.0 & 18.9 & 8.0 & 29.4 & 45.1 & 20.5 & 25.9 & 55.2 & 45.0 & 55.1 \\
mni & 8.9 & 4.0 & 0.4 & 13.2 & 23.6 & 9.7 & 14.9 & 17.7 & 36.2 & 47.5 \\
mup & 17.0 & 10.6 & 4.8 & 21.0 & 26.4 & 22.8 & 24.9 & 57.2 & 41.9 & 60.6 \\
mwr & 24.4 & 11.1 & 2.9 & 25.3 & 40.0 & 23.2 & 25.8 & 53.1 & 48.7 & 59.1 \\
sat & 1.3 & 0.3 & 0.0 & 2.6 & 26.6 & 6.7 & 16.7 & 6.2 & 27.5 & 38.6 \\
\midrule
\textbf{Avg.} & 16.5 & 13.5 & 7.0 & 23.4 & 37.4 & 16.3 & 24.8 & 49.0 & 46.0 & 55.9 \\
\textbf{High Avg.} & 17.7 & 18.9 & 10.8 & 25.8 & 41.9 & 16.4 & 29.5 & 64.7 & 49.4 & 57.3 \\
\textbf{Medium Avg.} & 13.5 & 8.3 & 2.8 & 21.6 & 36.7 & 14.0 & 23.9 & 45.8 & 50.0 & 57.9 \\
\textbf{Low Avg.} & 17.1 & 12.2 & 6.2 & 22.6 & 34.6 & 17.3 & 21.9 & 39.5 & 41.8 & 53.9 \\
\bottomrule
\end{tabular}
\caption{Performance comparison of models on \xorqaxx{} and \xorqa{} in one-shot setting across all supported languages.}
\label{tab:1shot_xorqa_detailed_results}
\end{table*}

\begin{table*}[h!]
\centering
\begin{tabular}{l@{}ccccc}
\toprule
& \multicolumn{5}{c}{\textbf{\xquad{}}} \\
\cmidrule(lr){2-6}
Lang. Code & \textbf{\llama{}-65B} & \textbf{\gemma{}-7B-IT} & \textbf{\bloomz{}-7B} & \textbf{\gpt{4}} & \textbf{\palm{}-L} \\
\midrule
en & 62.0 & 45.7 & 86.7 & 64.8 & 83.7 \\
bn & 7.4 & 35.7 & 57.1 & 56.5 & 72.3 \\
gu & 0.5 & 40.6 & 57.0 & 53.9 & 72.1 \\
hi & 25.3 & 49.4 & 63.7 & 63.1 & 76.7 \\
kn & 0.4 & 41.1 & 52.0 & 55.1 & 74.4 \\
ml & 3.6 & 33.7 & 48.8 & 56.5 & 66.6 \\
mr & 14.7 & 33.8 & 58.5 & 57.3 & 76.9 \\
ta & 4.0 & 42.1 & 55.0 & 55.3 & 75.1 \\
te & 0.2 & 36.8 & 51.9 & 48.8 & 68.0 \\
ur & 22.9 & 35.9 & 55.9 & 58.4 & 70.2 \\
as & 3.1 & 28.4 & 48.8 & 53.0 & 66.6 \\
or & 0.7 & 5.9 & 34.0 & 51.1 & 52.0 \\
pa & 1.8 & 40.2 & 61.4 & 59.7 & 60.7 \\
\midrule
\textbf{Avg.} & 7.1 & 35.3 & 53.7 & 55.7 & 69.3 \\
\textbf{High Avg.} & 8.8 & 38.8 & 55.5 & 56.1 & 72.5 \\
\textbf{Medium Avg.} & 1.9 & 24.8 & 48.1 & 54.6 & 59.8 \\
\bottomrule
\end{tabular}
\caption{Performance comparison of models on \xquad{} in one-shot setting across all supported languages.}
\label{tab:1shot_xquad_detailed_results}
\end{table*}

\begin{table}[h!]
\centering
\begin{tabular}{lcc}
\toprule
Lang. Code & \multicolumn{2}{c}{\textbf{\xquad{}}} \\
\cmidrule(lr){2-3}
& \textbf{\mt{} XXL} & \textbf{\palm{}-XS} \\
\midrule
en & 78.1 & 64.4 \\
bn & 53.2 & 47.7 \\
gu & 53.4 & 21.1 \\
hi & 59.2 & 59.1 \\
kn & 60.2 & 27.9 \\
ml & 52.6 & 30.1 \\
mr & 60.9 & 51.6 \\
ta & 62.8 & 45.1 \\
te & 51.0 & 28.2 \\
ur & 63.2 & 50.9 \\
as & 45.0 & 35.1 \\
or & 25.0 & 6.7 \\
pa & 51.3 & 8.8 \\
\midrule
\textbf{Avg.} & 53.2 & 34.4 \\
\textbf{High Avg.} & 57.4 & 40.2 \\
\textbf{Medium Avg.} & 40.5 & 16.9 \\
\bottomrule
\end{tabular}
\caption{Performance comparison of  \textbf{fine-tuned} models on \xquad{} dataset across all supported languages.}
\label{tab:ft_xquad_detailed_results}
\end{table}

\begin{table}[h!]
\centering
\begin{tabular}{@{}l@{}cc@{}}
\toprule
Lang. Code & \multicolumn{2}{c}{\textbf{\cssum{}}} \\
\cmidrule(lr){2-3}
& \textbf{\mt{} XXL} & \textbf{\palm{}-XS} \\
\midrule
en & 31.8 & 36.6 \\
bn & 25.4 & 28.8 \\
gu & 22.4 & 24.6 \\
hi & 26.2 & 30.5 \\
kn & 26.7 & 28.5 \\
ml & 26.4 & 26.7 \\
mr & 24.4 & 29.8 \\
ta & 31.6 & 32.3 \\
te & 25.8 & 26.9 \\
ur & 26.5 & 28.6 \\
as & 23.9 & 25.8 \\
bho & 22.0 & 24.1 \\
ne & 28.3 & 30.5 \\
or & 23.2 & 24.1 \\
ps & 26.9 & 24.9 \\
pa & 23.4 & 23.7 \\
sa & 25.4 & 26.0 \\
awa & 20.4 & 21.6 \\
brx & 19.7 & 12.6 \\
hne & 22.5 & 23.9 \\
gbm & 18.3 & 18.1 \\
bgc & 22.0 & 23.0 \\
gom & 25.1 & 25.8 \\
mai & 22.1 & 23.8 \\
mup & 24.5 & 25.6 \\
mni & 15.6 & 12.0 \\
mwr & 21.0 & 22.2 \\
hoj & 19.5 & 21.2 \\
sat & 5.9 & 10.0 \\
bo & 7.0 & 5.1 \\
\midrule
\textbf{Avg.} & 22.5 & 23.5 \\
\textbf{High Avg.} & 26.2 & 28.5 \\
\textbf{Medium Avg.} & 24.7 & 25.6 \\
\textbf{Low Avg.} & 18.7 & 18.8 \\
\bottomrule
\end{tabular}
\caption{Performance comparison of \textbf{fine-tuned} models on \cssum{} dataset across all supported languages.}
\label{tab:ft_cssum_detailed_results}
\end{table}

\begin{table*}[h!]
\centering
    \begin{tabular}{lcccccccccccccccccccccc}
    \toprule
    Lang. Code & \multicolumn{2}{c}{\textbf{\xorqaxx{}}} & & \multicolumn{2}{c}{\textbf{\xorqa{}}} \\
    \cmidrule{2-3} \cmidrule{5-6}
    & \textbf{\mt{} XXL} & \textbf{\palm{}-XS} & & \textbf{\mt{} XXL} & \textbf{\palm{}-XS} \\
    \midrule
    en & 72.4 & 55.8 & & 70.7 & 68.1 \\
    bn & 15.4 & 16.3 & & 71.9 & 68.6 \\
    gu & 28.7 & 25.9 & & 70.6 & 69.0 \\
    hi & 38.3 & 45.8 & & 70.6 & 68.5 \\
    kn & 28.2 & 32.0 & & 69.3 & 69.3 \\
    ml & 38.2 & 44.0 & & 71.1 & 70.5 \\
    mr & 32.8 & 32.9 & & 70.5 & 68.9 \\
    ta & 27.2 & 33.8 & & 69.0 & 69.8 \\
    te & 21.5 & 24.1 & & 70.5 & 70.4 \\
    ur & 28.3 & 18.6 & & 68.8 & 66.5 \\
    as & 24.8 & 27.6 & & 70.3 & 67.7 \\
    bho & 20.6 & 22.0 & & 67.9 & 66.0 \\
    or & 23.9 & 28.6 & & 69.1 & 65.3 \\
    pa & 27.3 & 28.5 & & 69.0 & 68.5 \\
    ps & 22.4 & 13.1 & & 68.3 & 64.2 \\
    sa & 22.1 & 21.5 & & 69.1 & 68.0 \\
    awa & 24.2 & 27.7 & & 68.0 & 65.2 \\
    bgc & 21.1 & 21.5 & & 69.1 & 62.5 \\
    bo & 41.5 & 6.4 & & 42.9 & 56.7 \\
    brx & 14.9 & 7.9 & & 38.8 & 30.2 \\
    gbm & 15.6 & 18.2 & & 65.2 & 62.5 \\
    gom & 17.8 & 21.0 & & 64.7 & 64.1 \\
    hne & 27.9 & 32.4 & & 68.2 & 64.7 \\
    hoj & 30.8 & 33.3 & & 66.2 & 62.1 \\
    mai & 23.5 & 31.8 & & 69.2 & 65.0 \\
    mni & 16.8 & 8.9 & & 48.6 & 37.4 \\
    mup & 20.1 & 19.1 & & 67.1 & 64.2 \\
    mwr & 27.2 & 23.7 & & 68.4 & 64.2 \\
    sat & 2.7 & 3.3 & & 32.1 & 36.5 \\
    \midrule
    \textbf{Avg.} & 24.5 & 23.9 & & 64.8 & 62.7 \\
    \textbf{High Avg.} & 28.8 & 30.4 & & 70.3 & 69.1 \\
    \textbf{Medium Avg.} & 23.6 & 23.6 & & 68.9 & 66.6 \\
    \textbf{Low Avg.} & 21.9 & 19.6 & & 59.1 & 56.6 \\
    \bottomrule
    \end{tabular}
\caption{Performance comparison of \textbf{fine-tuned} models on \xorqaxx{} and \xorqa{} dataset across all supported languages.}
\label{tab:ft_xorqa_detailed_results}
\end{table*}

\end{document}